%% file: main-arxiv.tex
\documentclass[10pt]{article}
\input{preamble.tex}

\usepackage{wrapfig}
\providecommand{\revnew}[1]{#1}
\usepackage{floatai-techreport}
\usepackage{enumitem}

\revisionmarksfalse

\newif\ifarxivmarks
\newif\ifpendingmarks
\arxivmarksfalse
\pendingmarksfalse
\definecolor{arxblue}{RGB}{0,40,175}
\definecolor{pendorange}{RGB}{200,90,0}
\newcommand{\arxnew}[1]{#1}
\newcommand{\arxlatest}[1]{#1}
\newcommand{\arxround}[1]{\ifarxivmarks{\color{arxblue}#1}\else#1\fi}

\newif\ifshowdraftnote
\showdraftnotefalse

\title{What Is a Repeated Token Worth? The Scaling Geometry of Multi-Epoch Pretraining}

\author[1,2]{Yekun Chai}
\author[3]{Haoyi Xiong}
\affiliation[1]{FloatAI}
\affiliation[2]{ETH Zurich}
\affiliation[3]{Independent Researcher}

\date{September 29, 2026}
\correspondence{\texttt{chaiyekun@gmail.com}}
\reportlogo{\includegraphics[height=0.65cm]{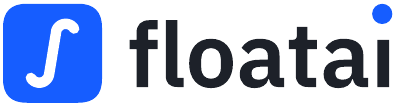}}

\begin{document}
\maketitle
\input{main-body-arxiv.tex}
\FloatBarrier
\bibliographystyle{iclr2027_conference}
\bibliography{references-arxiv}
\beginappendix
\input{appendix-arxiv.tex}
\end{document}

%% file: preamble.tex
\usepackage[utf8]{inputenc}
\usepackage[T1]{fontenc}
\usepackage{hyperref}
\usepackage{url}
\usepackage{booktabs}
\usepackage{amsfonts,amsmath,amssymb}
\usepackage{nicefrac}
\usepackage{microtype}
\usepackage{graphicx}
\usepackage{xcolor}
\usepackage{capt-of}
\usepackage{placeins}
\usepackage{longtable,array,pdfpages}
\definecolor{linkgreen}{RGB}{0,105,45}
\newif\ifrevisionmarks
\definecolor{reviseblue}{RGB}{0,40,175}
\newcommand{\revise}[1]{\ifrevisionmarks{\color{reviseblue}#1}\else#1\fi}
\hypersetup{colorlinks=true,breaklinks=true,linkcolor=linkgreen,citecolor=linkgreen,urlcolor=linkgreen}
\graphicspath{{figures/}{./}}

%% file: main-body-arxiv.tex
\begin{abstract}
As pretraining increasingly repeats data, every run faces three questions: how many epochs to take, how that
number should change with model size, and whether anything besides the epoch count matters.
We answer them by pricing a repeated token against two references: one epoch on the same data,
which gives its value, and fresh data at equal compute, which gives its cost. Against fresh data, the cost of repetition follows a single variable, the number
of extra epochs divided by the unique tokens per parameter. Against the same data, a second epoch is
worth nearly as much as a fresh one, and repeated tokens fall to half the value of fresh ones after a critical
epoch count that grows with the training budget per parameter but hardly with model size. With
unique data fixed, the predicted compute-optimal run grows model size and epochs together
until loss stops improving, near the critical epoch count. The same variable accounts for the direction of size trends that appear to conflict:
larger models tolerate fewer epochs when the corpus is fixed, from about 15 at 127M to 4 at 2B
parameters, but not when unique data grow with the model. Counts alone do not determine loss: at
identical counts, replaying shards consecutively raises loss by up to 0.46~bits per byte, concentrating
repeats on fewer samples also raises it, lower-entropy sources degrade faster with repetition, and re-tokenizing
repeats helps only under heavy repetition. These results offer an empirical guide to pretraining when unique data,
rather than compute, are the binding constraint.
\end{abstract}

\section{Introduction}
\label{sec:intro}
\arxnew{Compute-optimal scaling grows the training set with model size
\citep{kaplan2020scaling,hoffmann2022training}, but the stock of public human-generated text grows
more slowly \citep{villalobos2022run}, so pretraining increasingly repeats data. With a fixed pool
of $U$ unique tokens, \arxround{a model with $N$ parameters trained for $R$ epochs sees $D=RU$ training
tokens, and a run has to decide how many epochs to take}, how that number should change
for a larger model, and whether anything besides the epoch count matters. All three decisions turn
on one question: what is a repeated token worth?}

\begin{figure}[t]\centering
\includegraphics[width=\linewidth]{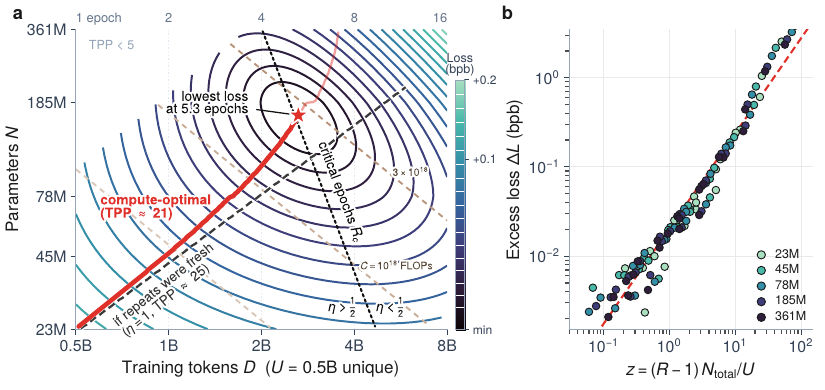}
\caption{\arxnew{\textbf{Excess loss grows with extra epochs and falls with unique data per parameter; with unique data fixed, the
predicted compute-optimal run \arxlatest{reaches its lowest loss near the point where a repeat is worth half a fresh token}.} CommonCrawl grid (\S\ref{sec:grid_setup}). (a)~Held-out loss at $U=0.5$B unique
tokens over training tokens $D=RU$ and model size $N$, from the one-epoch fit plus
Eq.~\eqref{eq:tpp_grid}, inside the grid's range (gray, training tokens per parameter, TPP $=D/N$, below 5); vertical lines mark epochs.
Red, the compute-optimal allocation (lowest loss at each compute $C=6ND$, \S\ref{sec:frontier});
light dashed, lines of equal compute; dark dashed, the same minimization with the one-epoch fit alone, as if every repeated token were worth a
fresh one ($\eta=1$); star, the lowest loss at this $U$\arxlatest{, beyond which more compute raises
loss (faded)}.
Dotted, the critical epoch count $R_c$ of Eq.~\eqref{eq:rc}, beyond which a repeated token is worth
less than half a fresh one ($\eta<\tfrac12$, Eq.~\ref{eq:ceg}). The optimum holds TPP \arxlatest{between 16 and 22}, below the fresh-data 25, and \arxlatest{reaches its lowest loss at 5.3 epochs, where $R_c\approx5.3$}. (b)~Excess loss over
fresh data at equal compute (Eq.~\ref{eq:excess_loss}) against \arxround{$z=(R-1)N_{\rm total}/U$} for the 113 of 140 repeated
configurations above the 0.002-bpb \arxround{(bits per byte)} resolution threshold; dashed, Eq.~\eqref{eq:tpp_grid}. \arxlatest{Per-size maps and critical epoch counts are in
Figure~\ref{fig:rc}.}}}
\label{fig:hero}\label{fig:tpp_grid}
\end{figure}

\arxnew{Prior work offers seemingly conflicting guidance. Up to about four epochs have been found
nearly as valuable as fresh data \citep{muennighoff2023scaling}, whereas repeating a subset degrades
models \citep{hernandez2022scaling}, and more so at larger sizes \citep{xue2023repeat}.
\citet{lovelace2026prescriptive} add a loss term in $(R-1)N/U$ and prescribe fewer epochs at large
compute, and \citet{qin2026bridging} find that the effectiveness of repeated tokens falls with tokens
per parameter. These studies differ along three methodological axes. First, a repeated run can be
compared with one epoch on the same unique data or with fresh data at equal compute. Second, across
model sizes, a study can hold $U$, $U/N$ or $D/N$ fixed. Third, the value of a repeat can be read off a
measured fresh-data loss curve or inferred by inverting a scaling law fitted to all runs, as in
effective-data formulations.}

\arxnew{We address all three. On a grid over model size, training tokens per parameter (TPP, $D/N$)
and epochs, every repeated run has both baselines, and every size has a measured one-epoch curve from
which the value of a repeat is read directly. We then test the resulting relations on designs that
hold $U$, $U/N$ or $D/N$ fixed, and finally hold all counts fixed to find what the counts miss.
\arxlatest{The results describe the cost of repetition as one surface over model size, unique data and
epochs, with a simple geometry: it collapses onto a single coordinate, $z$; a critical epoch count
separates repeats that keep most of their value from those that lose it; at fixed unique data, the
compute-optimal path ends near that boundary; and designs that hold $U$, $U/N$ or $D/N$ fixed are
different slices of the same surface, which is why they report different size trends.
Figure~\ref{fig:hero} shows this geometry at one $U$.} We find:}
\begin{enumerate}
\item \arxnew{\textbf{Excess loss grows with extra epochs and falls with unique data per parameter}
(\S\ref{sec:tpp_grid}). Relative to fresh data at equal compute, excess loss across all sizes and
budgets of the grid follows a power law in their ratio, $z=(R-1)N_{\rm total}/U$, \arxround{where $N_{\rm total}$ counts all
parameters including embeddings; no further size term is needed}.}
\item \arxnew{\textbf{Value, critical epochs and a predicted allocation} (\S\ref{sec:continuation},
\S\ref{sec:two_readings}, \S\ref{sec:frontier}). Relative to one epoch on the same data, a token of a
second epoch is worth 0.87--1.02 of a fresh token, and repeated tokens fall to half the value of
fresh ones at a \emph{critical epoch count} $R_c\approx2.7\,(D/N)^{0.24}$, nearly
independent of model size. At fixed unique data, the fitted compute-optimal run holds TPP slightly
below its fresh-data value and grows model size and epochs together \arxlatest{until loss stops
improving, near $R_c$}.}
\item \arxnew{\textbf{The size trend depends on what is held fixed} (\S\ref{sec:transfer}). At fixed $U$ the
loss minimum moves from about 15 to 4 epochs between 127M and 2B parameters; at fixed $U/N$ it
barely moves; at fixed $D/N$ larger models pay less for repetition. The direction of all three
trends, and of the reports above, matches how $z$ changes with size in each design
(Table~\ref{tab:reconcile}).}
\item \arxlatest{\textbf{Beyond counts} (\S\ref{sec:geometry}). The count variable carries over to
partial repetition, with a different exponent. At identical counts,
replaying shards consecutively, concentrating repeats on fewer samples and training on lower-entropy
text raise loss, and re-tokenizing repeats helps only under heavy repetition.}
\end{enumerate}

\begin{tcolorbox}[enhanced,breakable,colback=trBg,colframe=trLink!70!black,boxrule=0.6pt,arc=3pt,
  left=6pt,right=6pt,top=4pt,bottom=4pt,before skip=6pt,after skip=8pt,
  title={\textbf{Takeaways for pretraining on finite data}},fonttitle=\small,
  coltitle=white,colbacktitle=trLink!70!black]
\small
\begin{enumerate}[leftmargin=1.4em,itemsep=2pt,topsep=1pt]
\item \arxlatest{\textbf{How epochs should change with model size depends on the unique data.} With unique
data fixed, larger models should take fewer epochs; when unique data grow with the model, the best
epoch count barely changes (\S\ref{sec:transfer}).}
\item \arxlatest{\textbf{Larger training budgets per parameter tolerate more repetition.} The epoch count
at which repeated tokens fall to half the value of fresh ones grows with tokens per parameter and
hardly with model size (\S\ref{sec:two_readings}).}
\item \arxlatest{\textbf{With unique data fixed, grow model size and epochs together, up to about $R_c$.}
Additional compute goes to both in roughly equal shares until loss stops improving near the critical
epoch count (\S\ref{sec:frontier}).}
\item \arxlatest{\textbf{How tokens are repeated matters, not only how often.} Spread repeats across
epochs and samples, repeat lower-entropy sources less, and re-tokenize repeated data only under heavy
repetition (\S\ref{sec:geometry}).}
\end{enumerate}
\end{tcolorbox}

\section{Related Work}
\label{sec:related}
\input{related-work-arxiv.tex}

\begin{figure}[t]\centering
\includegraphics[width=\linewidth]{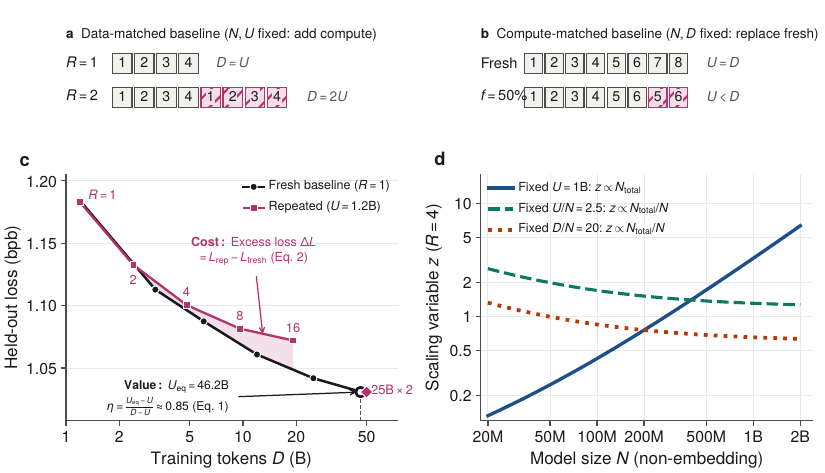}
\caption{\arxnew{\textbf{Study design and geometry of multi-epoch pretraining.} 
(a)~The data-matched baseline shares $N$ and $U$; extra epochs add training tokens $D=RU$ and compute. 
(b)~The compute-matched baseline shares $N$ and $D$; repeated data replace fresh data. 
\arxlatest{(c)~Both quantities on measured 59M losses: $\Delta L$ (Eq.~\ref{eq:excess_loss}) is the gap to the
compute-matched baseline at equal $D$, and $U_{\rm eq}$, the fresh budget with the same loss, gives
$\eta\approx0.85$ (Eq.~\ref{eq:ceg}). (d)~$z=(R-1)N_{\rm total}/U$ against model size at $R=4$ in the three
designs: with $U$ fixed, $z\propto N_{\rm total}$ (\S\ref{sec:fixed_u}); with $U/N$ (\S\ref{sec:un_slice}) or
$D/N$ (\S\ref{sec:dn_slice}) fixed, $z\propto N_{\rm total}/N$, which falls slowly as the embedding share
shrinks.}}}
\label{fig:comparisons}
\end{figure}

\section{\texorpdfstring{\arxnew{Measuring Repetition}}{Measuring Repetition}}
\label{sec:comparisons}\label{sec:design}
\arxnew{\paragraph{Value and cost of a repeated token.}
A run of $R$ epochs over $U$ unique tokens trains on $D=RU$ tokens. We compare it with two
one-epoch baselines of the same model size $N$ (Figure~\ref{fig:comparisons}). The
\emph{data-matched baseline} trains once on the same $U$ tokens. The repeated run spends $(R-1)U$
more tokens of compute, and the comparison asks whether a data-constrained run should keep training.
We express the gain in fresh tokens: let $U_{\rm eq}$ be the fresh budget whose one-epoch loss equals
that of the repeated run, the same inversion that defines effective data transferred in
\citet{hernandez2021transfer}. The \emph{value of a repeated token}, which we call the
\emph{fresh-token equivalence} $\eta$, is}
\begin{equation}
\eta=\frac{U_{\rm eq}-U}{D-U},\label{eq:ceg}
\end{equation}
\arxnew{the fresh-equivalent tokens gained per repeated token: $\eta=1$ if repeats were as good as
fresh data and $\eta=0$ if they were worthless. Because $\eta$ averages over all $D-U$ repeated
tokens, a value that falls with $R$ means that the last epoch is worth less than the average. In
effective-data forms, $\eta$ is the discount on repeated tokens \citep{muennighoff2023scaling}. The \emph{compute-matched baseline} trains once on $D$
fresh tokens, so repeated tokens replace fresh ones at equal compute and schedule. The
\emph{excess loss}
\begin{equation}
\Delta L = L(N, U, R) - L(N, D, 1)
\label{eq:excess_loss}
\end{equation}
\arxround{where $L(N,U,R)$ is the held-out loss of an $N$-parameter model after $R$ epochs over $U$ unique tokens,}
is the direct cost of repetition at matched parameters and compute. In \emph{full-corpus repetition}
every sample is seen $R$ times. In \emph{partial repetition}, a fraction $f$ of the training targets
comes from a subset seen $R_{\rm sub}$ times and is mixed uniformly with fresh data, so
$U=D(1-f+f/R_{\rm sub})$. Throughout, $D$ and $U$ count prediction targets, $D$ including repeats.}

\paragraph{\arxnew{Models, data and evaluation.}}
\label{sec:sweep_protocol}\label{sec:tpp_definition}\label{sec:measurement}\label{sec:uncertainty}
\arxnew{All models are decoder-only Transformers trained with Muon
\citep{jordan2024muon}\footnote{\url{https://github.com/KellerJordan/Muon}} for hidden matrices and
Adam \citep{kingma2015adam} elsewhere, with the learning rate annealed to each run's final horizon
(Appendix~\ref{app:protocols}); model sizes are non-embedding parameter counts $N$, except in the
fixed-$U$ sweeps (\S\ref{sec:fixed_u}), which use a separate \arxlatest{\emph{proxy family}} whose sizes are
$12Ld^2$ labels \arxround{for $L$ layers of width $d$} (Appendix Table~\ref{tab:ladder}). All runs use
SlimPajama \citep{soboleva2023slimpajama}, a cleaned and deduplicated version of RedPajama, tokenized
with the GPT-2 byte-level BPE. Single-source runs train and evaluate on one of its seven subsets;
\emph{mixed} sweeps use all seven in their natural proportions. We report bits per byte (bpb) on
held-out test sets of each source, after removing training windows that exactly match a test window;
recipes and repeat counts are chosen on separate development sets. Each experiment holds a different
combination of $N$, $U$, $D$ and $R$ fixed; we describe each where it is used, and Appendix
Table~\ref{tab:families} summarizes them.}

\section{\texorpdfstring{\arxnew{The Landscape of Repetition}}{The Landscape of Repetition}}
\label{sec:map}\label{sec:grid_setup}
\arxnew{We map repetition on a CommonCrawl grid over model size \arxlatest{(23M to 361M)}, training budget (5 to 300 TPP) and
epoch count (1 to 16). Within each size and budget, runs differ only in
their data, and each size has a measured one-epoch curve that supplies both baselines
(Appendix~\ref{app:grid}). From the grid we read the value of a repeated token
(\S\ref{sec:continuation}), its cost (\S\ref{sec:tpp_grid}), the critical epoch count that links the
two (\S\ref{sec:two_readings}) and the allocation of a fixed pool of unique data
(\S\ref{sec:frontier}).}

\subsection{\texorpdfstring{\arxnew{Value of a repeated token}}{Value of a repeated token}}
\label{sec:continuation}\label{sec:ds}
\arxnew{We obtain $U_{\rm eq}$ by interpolating each model size's measured one-epoch curve
(Appendix~\ref{app:grid}).}

\begin{wrapfigure}[19]{r}{0.40\linewidth}
\vspace{-1.2em}\centering
\includegraphics[width=\linewidth]{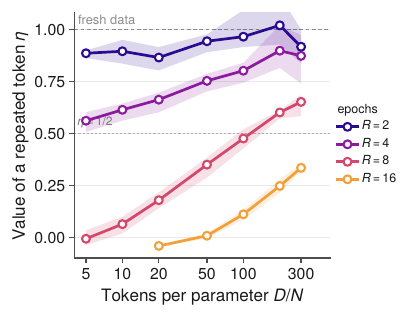}
\vspace{-1.6em}
\caption{\arxnew{\textbf{Repeats lose value with epochs, fastest at small budgets.} $\eta$ (Eq.~\ref{eq:ceg}):
mean (lines) and range (shading) over five sizes. \arxlatest{Sixteen-epoch runs at $D/N\le20$ are omitted: their loss lies
above the shortest one-epoch run, so $U_{\rm eq}$ cannot be read off.}}}
\label{fig:fte}
\vspace{-1.0em}
\end{wrapfigure}
\indent\arxnew{A repeated token loses value with every further epoch, and loses it fastest at small
budgets per parameter (Figure~\ref{fig:fte}). The value falls with $R$ at every budget and broadly
rises with TPP, but the gap between epoch counts narrows as TPP grows: from two to eight epochs it
drops from about 0.9 to zero \arxlatest{or below} at $D/N=5$, yet only from about 0.9 to 0.65 at 300. Only the second epoch
is nearly as good as fresh data at every budget (0.87--1.02). Model size matters less: at $R\ge8$ the
five sizes lie within 0.10 of each other at every budget. The rise with TPP is opposite to the
decline that \citet{qin2026bridging} infer from a fitted law; \arxlatest{\S\ref{sec:tpp_grid} discusses why}.}

\indent\arxnew{\arxlatest{Separate sweeps that train eight sizes on the same 1B unique tokens (\S\ref{sec:fixed_u})} agree in direction. Among their models with one-epoch references,
$U/N$ at $U=1$B falls 17-fold from 59M to 1B; a second epoch keeps its value (0.93--1.01) while four epochs fall from 0.81 to 0.59
(Figure~\ref{fig:rpeak}c), and their two-epoch level is close to the grid's (Appendix
Table~\ref{tab:second_epoch}). On five sources at 59M, a second epoch
likewise recovers 98--101\% of the gain from fresh data (Appendix~\ref{app:model_checks}).}

\subsection{\texorpdfstring{\arxnew{Excess loss tracks extra epochs per unique token per parameter}}{Excess loss tracks extra epochs per unique token per parameter}}
\label{sec:tpp_grid}
\arxnew{\arxlatest{A single quantity sets the cost of repetition at every size and budget \arxlatest{of the grid}: the number of extra
epochs per unique token per parameter.} \arxlatest{Across the repeated runs}, excess loss over the
compute-matched baseline follows}
\begin{equation}
\Delta L = A\,z^{p},\qquad z=(R-1)\frac{N_{\rm total}}{U}=\frac{(R-1)R}{D/N}\cdot\frac{N_{\rm total}}{N},
\label{eq:tpp_grid}
\end{equation}
\arxnew{with $A=0.0203$~bpb and $p=1.07$ ($R^2=0.96$; Figure~\ref{fig:hero}b). $z$ is the number of extra
epochs divided by the unique tokens per parameter, as in the repetition term of
\citet{lovelace2026prescriptive}: a repeat costs more the more often each token is seen and the less
unique data each parameter has. With $N_{\rm total}$ the parameter count including embeddings, a
further factor $N_{\rm total}^{-g}$ is negligible ($g=-0.02$, about 4\% across the grid); counting
only non-embedding parameters leaves a residual size dependence, which we attribute to embeddings
making up 53\% of the 23M model but only 15\% of the 361M model (Appendix
Figure~\ref{fig:grid_collapse}). \arxlatest{How large $g$ is depends on the setup (Appendix~\ref{app:fixed_budget}).}}

\arxnew{\paragraph{Why measured curves.} We read $U_{\rm eq}$ off measured one-epoch curves rather than
a fitted law, for two reasons. First, effective-data laws cannot represent the largest repetition
costs. \arxlatest{Such laws \citep{muennighoff2023scaling,qin2026bridging} treat} repeated training as fresh training on a
discounted amount of data,
\[
L=E+A'N_{\rm total}^{-\alpha}+B\,\big[U+\eta(D-U)\big]^{-\beta},\qquad 0\le\eta\le1,
\]
\arxround{with irreducible loss $E$ and fitted constants $A'$, $B$, $\alpha$ and $\beta$,} so no repeated run can end above the loss of one epoch over its $U$ unique tokens. About a quarter of
the repeated runs that have a one-epoch run at their $U$ do end above it, nearly all at the largest
$z$. Adding a repetition cost in $z$ to a Chinchilla backbone instead,
\[
L=E+A'N_{\rm total}^{-\alpha}+B\,D^{-\beta}+\arxround{G}\,z^{p'},
\]
\arxround{with a fitted coefficient $G$ and exponent $p'$,} removes this bound and fits all 175 grid runs to 3\% RMS in $\log L$, against 13\% for the
effective-data form (Appendix~\ref{app:grid}). Second, inverting any fitted law for $\eta$ amplifies
small loss residuals: where the one-epoch curve is flat, a residual of a few thousandths of a bpb
shifts the inferred effective data substantially. On the grid, values inverted this way show no
consistent trend with TPP, whereas those read off measured curves rise with it
(Appendix~\ref{app:qin}).}

\subsection{\texorpdfstring{\arxnew{The critical epoch count}}{The critical epoch count}}
\label{sec:two_readings}\label{sec:rc}
\begin{figure}[t]\centering
\includegraphics[width=\linewidth]{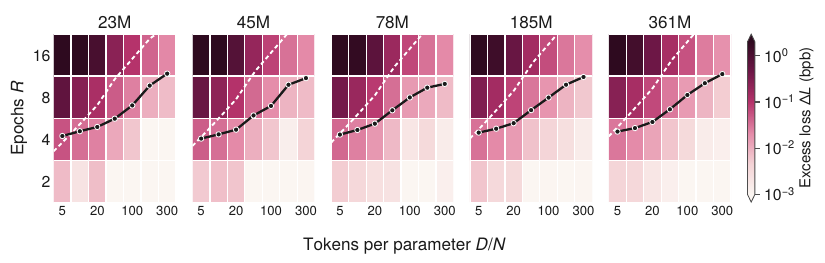}
\caption{\arxlatest{\textbf{Repeated tokens fall to half the value of fresh ones at a critical epoch count that
grows with the training budget and hardly with model size.} Excess loss over fresh data at equal
compute on the CommonCrawl grid, for every size, budget and epoch count (values in Appendix
Figure~\ref{fig:map}). Black points, the measured $R_c$, where $\eta=\tfrac12$, interpolated in $\log R$;
white dashed, the 0.1-bpb level of Eq.~\eqref{eq:tpp_grid}.}}
\label{fig:rc}
\end{figure}
\arxlatest{Value and cost combine into a single epoch count per budget. Since $\eta$ falls with $R$ at
every budget, we define the \emph{critical epoch count} $R_c$ as the number of epochs at which
$\eta=\tfrac12$, where repeated tokens are on average worth half a fresh one\arxlatest{, a measured counterpart of the
constant ``half-life'' of repeated data in the fitted law of \citet{muennighoff2023scaling}}; unlike a loss tolerance,
this threshold is dimensionless, so it compares budgets whose losses differ in scale. $R_c$ is about four to five epochs up to a TPP of 20, rises to
about ten at 300 and barely depends on model size (Figure~\ref{fig:rc}). The measured
crossings follow \arxlatest{(Appendix Figure~\ref{fig:map}c)}}
\begin{equation}
R_c\approx2.7\,(D/N)^{0.24}.
\label{eq:rc}
\end{equation}
\arxlatest{$R_c$ rises with the budget because value and cost scale differently: one-epoch loss falls with
data as $D^{-\beta}$, whereas excess loss falls with unique data per parameter as $(U/N)^{-p}$ with
$p>\beta$, so the cost of a repeat shrinks faster than the fresh-data gain it forgoes. \arxlatest{Because the fresh-data gain itself shrinks with the budget, $R_c$ grows more slowly than the $\sqrt{D/N}$
that a fixed loss tolerance would imply} through Eq.~\eqref{eq:tpp_grid}. $R_c$ is defined by the value of a repeat, not by an allocation;
\S\ref{sec:frontier} finds that the compute-optimal allocation at fixed $U$ stops improving near it.}

\subsection{\texorpdfstring{\arxnew{Compute-optimal allocation at fixed unique data}}{Compute-optimal allocation at fixed unique data}}
\label{sec:frontier}
\begin{figure}[t]\centering
\includegraphics[width=\linewidth]{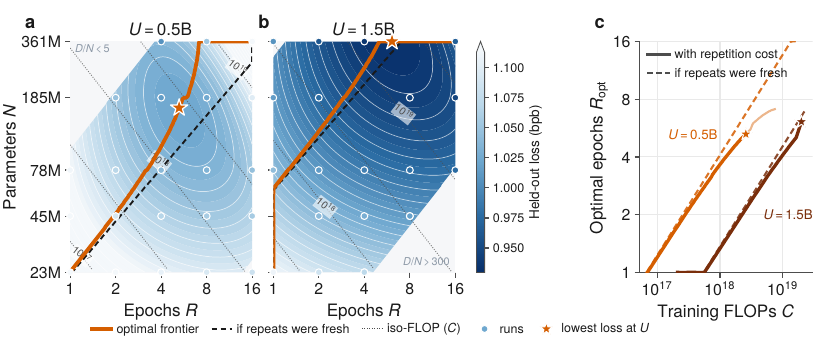}
\caption{\arxnew{\textbf{\arxlatest{At fixed unique data, the fitted compute-optimal frontier grows
model size and epochs together until loss stops improving.}} (a,b)~Predicted loss over epochs $R$ and model
size $N$ at $U=0.5$B and 1.5B (fresh-data fit plus Eq.~\ref{eq:tpp_grid}), within the grid's range
($1\le R\le16$, $5\le D/N\le300$); since $C=6NRU$, the dotted diagonals are iso-FLOP lines, labeled in
FLOPs. The solid line minimizes loss along each of them; the dashed line ignores the cost of repetition.
Points are grid runs with $R\ge2$ and $U$ within a factor 1.41 of the panel's, colored by their loss
shifted to this $U$ by the fit. \arxlatest{Stars mark the lowest predicted loss at each $U$, at the
grid's largest model for 1.5B; beyond it, more compute raises loss.} (c)~Epochs on the frontier;
\arxlatest{fitted up to the stars, $R_{\rm opt}\propto C^{0.47}$ at $U=0.5$B and $C^{0.48}$ at 1.5B;
faded beyond them}.}}
\label{fig:frontier}
\end{figure}
\arxnew{Given $U$ unique tokens, a run with compute $C=6NRU$ must split it between model size and
epochs. We predict loss over $(N,R)$ from a Chinchilla fit to the one-epoch runs plus
Eq.~\eqref{eq:tpp_grid} and minimize it at each $C$ (Figures~\ref{fig:hero}a and~\ref{fig:frontier};
fitted and measured iso-loss contours are compared in Appendix Figure~\ref{fig:gridisoloss}).}

\arxlatest{With unique data fixed, the compute-optimal run grows model size and epochs together, each
roughly as $\sqrt{C}$, until loss stops improving at an epoch count close to $R_c$.

\indent Size and epochs grow together because the frontier holds TPP nearly fixed. Up to the loss
minimum, TPP stays at \arxlatest{16--24}, at or below the fresh-data optimum of the same fit (\arxlatest{about 25}): repeated tokens
are worth less than fresh ones, so the optimum moves compute from data to parameters. With $D/N$ fixed
at $\tau$, $C=6\tau N^2$ gives
\begin{equation}
N_{\rm opt}=\sqrt{C/(6\tau)},\qquad R_{\rm opt}=\tau N_{\rm opt}/U=\sqrt{\tau C/6}\,\big/\,U .
\label{eq:frontier}
\end{equation}
The fitted exponents, $N_{\rm opt}\propto C^{0.52\text{--}0.53}$ and
$R_{\rm opt}\propto C^{0.47\text{--}0.48}$, are close to these square roots; $R_{\rm opt}$ grows
slightly more slowly because TPP drifts down along the frontier (Appendix~\ref{app:grid}). \arxlatest{The fitted
law of \citet{muennighoff2023scaling} instead scales epochs slightly faster than parameters; both
agree that compute at fixed $U$ should go to both.}

\indent Loss stops improving at a finite compute. The lowest predicted loss lies at 164M parameters and 5.3
epochs for $U=0.5$B and at \arxlatest{297M} and 5.6 epochs for 1B, against $R_c\approx5.3$ and 5.5 at the same TPP;
for 1.5B it reaches the grid's largest model, at 6.1 epochs against $R_c\approx5.9$
(stars in Figure~\ref{fig:frontier}). Beyond this point, any split of additional compute within the grid
raises loss: a larger model and more epochs both raise $z$, and the cost outgrows the gain. A fixed pool of unique
data therefore bounds both the useful model size and the useful epoch count. The proximity to $R_c$ is
empirical: $R_c$ concerns the average value of repeated tokens, \arxlatest{whereas the loss minimum marks where
additional compute stops paying off}.

\indent The allocation is insensitive to the cost of repetition. Minimizing the one-epoch fit alone,
as if repeats were fresh, picks models up to a fifth smaller, trained for more epochs, yet loses at
most 0.002~bpb up to the loss minimum, because the loss surface is flat near the frontier.

\indent These conclusions describe the fitted loss surface over the grid's sizes and budgets. Past the
minimum, further gains at fixed $U$ would come from models larger than the grid trained below a TPP of
5, a regime that \citet{lovelace2026prescriptive} model with separate exponents for epochs and the
parameter-to-data ratio; there, additional compute favors larger models and fewer epochs, consistent
with the faster growth of model size in our fixed-$U$ sweeps (Appendix~\ref{app:compute}). \arxlatest{The minimum
itself depends on the recipe: with strong weight decay and ensembling, loss can keep falling with model
size at fixed data \citep{kim2025infinite}.}}

\section{\texorpdfstring{\arxnew{Size Trends Depend on What Is Held Fixed}}{Size Trends Depend on What Is Held Fixed}}
\label{sec:transfer}\label{sec:scale}\label{sec:three_designs}
\arxnew{In Eq.~\eqref{eq:tpp_grid}, model size enters only through $z=(R-1)N_{\rm total}/U$. How the cost
of repetition changes with size therefore depends on what a design holds fixed: with $U$ fixed, $z$
grows in proportion to $N_{\rm total}$; with $U/N$ or $D/N$ fixed at a given $R$, it changes only
through $N_{\rm total}/N$, which falls as the embedding share shrinks (from 2.1 at 23M to 1.08 at
1.2B). We test each prediction on a design outside the grid.}

\subsection{\texorpdfstring{\arxnew{Fixed unique data: larger models reach the minimum sooner}}{Fixed unique data: larger models reach the minimum sooner}}
\label{sec:rpeak}\label{sec:fixed_u}
\begin{figure}[t]\centering
\includegraphics[width=\linewidth]{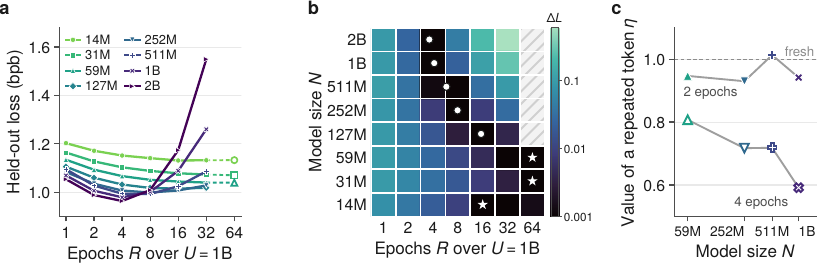}
\caption{\arxnew{\textbf{At fixed $U$, larger models reach their loss minimum after fewer epochs,
while a second epoch keeps its value.} CommonCrawl, $U=1$B, separately annealed runs. (a)~Held-out
loss; \arxlatest{open markers, additional runs at $R=64$ (Appendix~\ref{app:r64})}. (b)~The full sweep: loss above each model's lowest measured loss, $\Delta L$ (bpb), for every model size
and epoch count (hatched, not run). White dots, parabola minima in $\log R$ for the five sizes whose loss turns
upward; stars, lowest measured loss for the three smallest models (14M, 31M, 59M), which have not
turned upward by $R=64$. (c)~$\eta$ at $R=2$ (filled) and $R=4$ (open),
inverting a power law through each size's one-epoch endpoints at $U=1$B and 4B (Appendix
Table~\ref{tab:second_epoch}).}}
\label{fig:rpeak}\label{fig:rstarcollapse}
\end{figure}
\arxnew{\arxlatest{When unique data are fixed, larger models reach their loss minimum after fewer
epochs and degrade faster past it.} \arxlatest{We train eight sizes from 14M to 2B on
the same 1B unique CommonCrawl tokens (the \emph{fixed-$U$ sweeps}), each to its own horizon.} For the five sizes from 127M to 2B, loss
turns upward within 32 epochs, with minima at
15.3, 7.9, 5.8, 4.1 and 3.6 epochs (Figure~\ref{fig:rpeak}a,b), \arxlatest{summarized over this range by
$R^\ast\approx4.5\,(U/N)^{0.51}$\arxround{, where $R^\ast$ is the loss-minimizing epoch count}. Past the minimum, larger models also degrade faster: only the 1B and
2B models end worse than one epoch at 32 epochs, while the 14M, 31M and 59M models \arxlatest{show no clear upturn by 64 epochs}.} Relative to the data-matched baseline, a second epoch keeps its value at every size while four
epochs lose part of theirs as size grows (Figure~\ref{fig:rpeak}c), the direction of the $U/N$ trend
of \S\ref{sec:continuation}.}

\begin{figure}[t]\centering
\includegraphics[width=\linewidth]{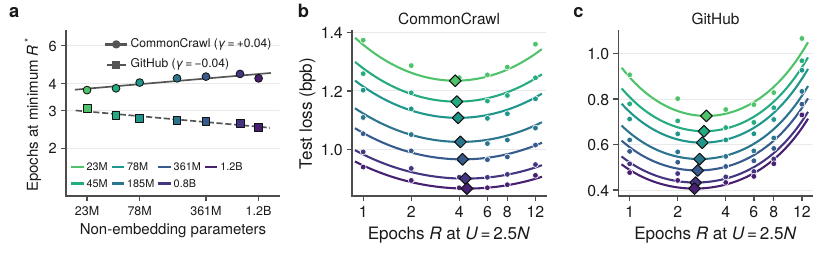}
\caption{\revnew{\textbf{When unique data grow with model size, the loss-minimizing epoch count barely
changes.} Full-corpus repetition at $U=2.5N$ (so $D/N=2.5R$) for seven exact non-embedding sizes from
23M to 1.2B; test loss of separately annealed runs. (a)~Epochs at the loss minimum,
from a least-squares quadratic in $\log R$ through all points of each size, and power laws
$R^*\propto N^{\gamma}$ through them. (b,c)~Test loss (points) and a six-parameter surface fitted
per source ($\log L$ quadratic in $\log R$, with coefficients linear in $\log N$); diamonds mark the
surface minimum for each size.}}
\label{fig:scale}\label{fig:fixed_pool}
\end{figure}

\subsection{\texorpdfstring{\arxnew{Fixed unique tokens per parameter: the minimum is stable}}{Fixed unique tokens per parameter: the minimum is stable}}
\label{sec:fixed_pool}\label{sec:un_slice}
\arxnew{\arxlatest{When unique data grow in proportion to model size, the loss-minimizing epoch count
barely moves.} We give each of seven sizes from 23M to 1.2B the same unique budget per parameter, $U=2.5N$,
and train it for $R\in\{1,2,4,6,8,12\}$ epochs on CommonCrawl and on GitHub
(Figure~\ref{fig:scale}). On CommonCrawl, four or six epochs give the lowest measured loss at every
size, and the fitted minimum stays between 3.7 and 4.4 epochs; on GitHub it drifts from 3.1 to 2.5
epochs. Both trends are far flatter than at fixed $U$, where the minimum falls fourfold over a 16-fold
range of sizes. Past the minimum the sources diverge: on
CommonCrawl, 12 epochs still beat one at every size, by 0.014--0.043~bpb, whereas on GitHub they
are worse by 0.16--0.25~bpb, in line with its lower entropy (\S\ref{sec:source}).}

\subsection{\texorpdfstring{\arxnew{Fixed TPP: larger models pay less}}{Fixed TPP: larger models pay less}}
\label{sec:matched_tpp}\label{sec:dn_slice}
\begin{wrapfigure}[17]{r}{0.40\linewidth}
\vspace{-1.6em}\centering
\includegraphics[width=\linewidth]{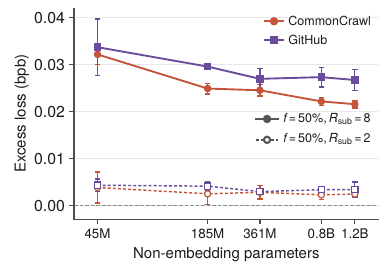}
\vspace{-1.9em}
\caption{\arxnew{\textbf{At $D=20N$, the cost of heavy partial repetition shrinks with size but persists.}
Excess loss with half of the budget from a subset repeated $R_{\rm sub}=8$ (filled) or 2 (open) times;
means with 95\% $t$ intervals over three paired replicates.}}
\label{fig:fraction_scale}
\vspace{-1.4em}
\end{wrapfigure}
\indent\arxnew{\arxlatest{At a fixed training budget per parameter, larger models pay less for the same
repetition, though the cost does not vanish.} \arxlatest{We train five sizes from 45M to 1.2B at a fixed budget per parameter, $D=20N$, on
CommonCrawl and GitHub, either on fresh data or with half of the budget drawn from a subset repeated
$R_{\rm sub}=2$ or 8 times (Figure~\ref{fig:fraction_scale}). Repeating the subset eight times} costs 0.032 and 0.034~bpb (CommonCrawl and
GitHub) at 45M, and 0.022 and 0.027~bpb at 1.2B: a third and a fifth smaller, but positive at every
size, with intervals that exclude zero. Repeating the subset twice costs at most 0.004~bpb at any
size. At fixed $D/N$, the subset's $z$ changes with size only through $N_{\rm total}/N$, which falls
from 1.71 at 45M to 1.08 at 1.2B; with exponents between those of partial repetition (0.53,
\S\ref{sec:surface}) and the grid (1.07), this predicts 22--39\% less excess loss at 1.2B,
against the measured 33\% and 21\%. \arxlatest{A separate comparison of 45M and 185M models at $D\approx4.5N$, with three paired
replicates per source,} gives the same sign: 21\% less at 185M than at 45M (Appendix~\ref{app:paired_scale}).}

\begin{table}[t]\centering\footnotesize\setlength{\tabcolsep}{4pt}
\caption{\arxnew{\textbf{The direction of each size trend matches how $z$ changes with model size.}
$z=(R-1)N_{\rm total}/U$; $N_{\rm total}/N$ falls from 2.1 at 23M to 1.08 at 1.2B. Predictions for the
published results use Eq.~\eqref{eq:tpp_grid}.}}
\label{tab:reconcile}
\newcolumntype{L}[1]{>{\raggedright\arraybackslash}p{#1}}
\begin{tabular}{@{}L{3.5cm}L{2.1cm}L{1.6cm}L{2.9cm}L{3.3cm}@{}}
\toprule
Design & Baseline & $z$ with size & Predicted & Observed\\
\midrule
Fixed $U$ (\S\ref{sec:rpeak}) & data-matched & $\propto N_{\rm total}$ & minimum at fewer epochs & 15.3 to 3.6 epochs, 127M to 2B\\
Fixed $U/N$ (\S\ref{sec:fixed_pool}) & data-matched & $\propto N_{\rm total}/N$ & minimum nearly fixed & \leavevmode\arxnew{3.7--4.4} epochs on CommonCrawl\\
Fixed $D/N$, $f$, $R_{\rm sub}$ (\S\ref{sec:matched_tpp}) & compute-matched & $\propto N_{\rm total}/N$ & \arxnew{22--39\%} less at 1.2B & 33\% and 21\% less\\
\midrule
Fixed $U$, \arxlatest{T5} \citep{xue2023repeat} & compute-matched & $\propto N_{\rm total}$ & larger models degrade more & larger models degrade more\\
$R=4$, $D\approx20N$ \citep{muennighoff2023scaling} & compute-matched & $\approx0.6$ & \arxnew{$\Delta L\approx0.01$~bpb} & nearly as good as fresh data\\
Small subset, many repeats \citep{hernandez2022scaling} & compute-matched & large & large excess loss & marked degradation\\
\bottomrule
\end{tabular}
\end{table}

\subsection{\texorpdfstring{\arxlatest{How $z$ scales with size explains the opposing trends}}{How z scales with size explains the opposing trends}}
\label{sec:reconcile}

\arxlatest{The three designs and the published results all follow from how $z$ scales with model size and
which baseline the cost is measured against (Table~\ref{tab:reconcile}).} \arxnew{When $U$ is fixed, $z$
grows with size; when $U/N$ or $D/N$ is fixed, it shrinks slightly. The fixed-$U/N$ design reads the loss
minimum against the data-matched baseline, whereas the fixed-$D/N$ design reads excess loss against the
compute-matched one. The cost of repetition is therefore not a function of TPP alone.} \arxlatest{The same
accounting covers the studies cited in \S\ref{sec:intro}.} \arxnew{Near compute-optimal budgets, four epochs over
$U=D/4$ with $D\approx20N$ give $z\approx0.6$ and an excess loss of about 0.01~bpb by
Eq.~\eqref{eq:tpp_grid}, consistent with the finding that a few epochs at equal compute are nearly as good
as fresh data \citep{muennighoff2023scaling}. A small subset repeated many times has a large $z$ \arxround{computed on the subset ($z_{\rm sub}$, \S\ref{sec:surface})}
and degrades the model \citep{hernandez2022scaling}, and at fixed $U$ the cost grows with size because $z$
does \citep{xue2023repeat}.} \arxlatest{Comparing studies therefore requires matching the baseline and the
fixed quantity, not only token counts. In practice, a larger model needs fewer epochs only when it is given
the same unique data.}

\section{\texorpdfstring{\arxnew{Beyond Token Counts}}{Beyond Token Counts}}
\label{sec:geometry}\label{sec:interventions}
\arxlatest{Counts set most of the cost of repetition, but not all of it. Repeating only part of the data
follows a different law from repeating the whole corpus (\S\ref{sec:surface}); at identical counts, the
order, allocation and source of the repeats change loss (\S\ref{sec:schedule}); and re-tokenizing
repeats helps only under heavy reuse (\S\ref{sec:bonus}).}

\subsection{\texorpdfstring{\arxlatest{Partial repetition}}{Partial repetition}}
\label{sec:surface}\label{sec:scaling}\label{sec:rstar}\label{sec:joint}\label{sec:replacement}\label{sec:iso}
\arxlatest{In practice, a training budget is often filled partly with repeated data rather than by repeating
the whole corpus. At a fixed budget, partial repetition does not measurably improve on fresh data. We train 45M models at
$D\approx18N$ on CommonCrawl, GitHub and Books, drawing a fraction $f$ of the training tokens from a subset
seen $R_{\rm sub}$ times, for nine allocations $(f,R_{\rm sub})$ with three paired replicates each. Every
allocation is worse than fresh data on every source, and the cost rises with $R_{\rm sub}$ at fixed $f$
(Appendix Figure~\ref{fig:replacement}a--c); broader sweeps \arxlatest{from 59M to 2B} agree
(Appendix~\ref{app:fixed_budget}).}

\arxlatest{The cost is, however, smaller than full-corpus repetition implies. With the subset's own variable
$z_{\rm sub}=(R_{\rm sub}-1)N_{\rm total}/(fD/R_{\rm sub})$, Eq.~\eqref{eq:tpp_grid} scaled by $f$
overstates the measured costs about fourfold, whereas a free fit gives
$\Delta L\propto f^{1.63}z_{\rm sub}^{0.53}$ ($R^2=0.89$; Appendix~\ref{app:partial}). The cost grows faster
than linearly in the repeated fraction but more slowly in $z_{\rm sub}$ than under full-corpus
repetition, so a small subset repeated many times is cheaper than its $z_{\rm sub}$ suggests, and
Eq.~\eqref{eq:tpp_grid} serves as a conservative estimate. \arxlatest{These fits cover up to 32 repeats of the subset;
at far higher counts, the damage peaks at intermediate repeat counts
\citep{hernandez2022scaling,chudnovsky2026fr}.} With size, the cost shrinks at fixed $D/N$
but persists (\S\ref{sec:matched_tpp}).}

\subsection{\texorpdfstring{\arxnew{Order, allocation and source at fixed counts}}{Order, allocation and source at fixed counts}}
\label{sec:schedule}\label{sec:prescriptive}
\begin{figure}[t]\centering
\includegraphics[width=\linewidth]{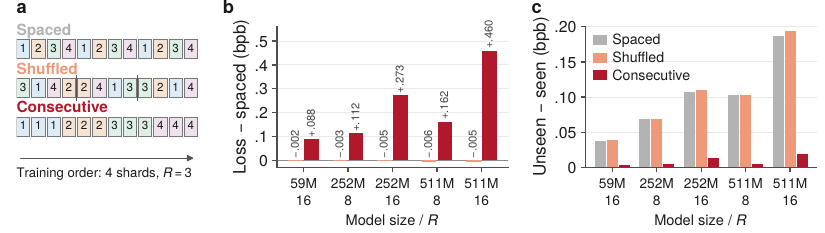}
\caption{\textbf{Large loss increases appear under consecutive repetition, not \revise{per-epoch}
shuffling, at fixed $(N,U,R)$.} Sweeps on CommonCrawl at $U=1$B. (a)~Schematic of the three orders of the same data, drawn at shard level. (b)~Loss of shuffled
and consecutive runs minus that of the spaced run with the same $(N,U,R)$.
(c)~Unseen-minus-seen loss gap: consecutive order nearly removes the gap while
raising loss, so a smaller gap does not by itself indicate better generalization.}
\label{fig:schedule}
\end{figure}

\arxlatest{At fixed $(N,U,R)$, the order, allocation and source of repeats still change loss.}
\arxnew{\paragraph{Order.}
\arxlatest{Repeats should be spaced apart: replaying each shard many times in a row is far more costly than
revisiting it once per epoch, while the order of shards within an epoch hardly matters\arxlatest{, in line with
replay schedules that space repeats \citep{zuo2026train}}. For models from
59M to 511M on 1B unique CommonCrawl tokens, we present the same tokens in three orders
(Figure~\ref{fig:schedule}a): \emph{spaced} runs $R$ successive epochs (the default \arxlatest{in these sweeps}), \emph{shuffled}
reorders the shards of each epoch, and \emph{consecutive} repeats each shard $R$ times in a
row. Consecutive replay raises held-out loss by up to 0.46~bpb, whereas shuffling changes it by less
than 0.01~bpb (Figure~\ref{fig:schedule}b). Consecutive replay also nearly closes the \emph{unseen-minus-seen gap}, the difference between loss on
held-out text and on text already trained on (Figure~\ref{fig:schedule}c), so a small gap need not imply
low held-out loss.}}

\arxnew{\paragraph{Allocation.}\label{sec:matched_content}
\arxlatest{Repeats should also be spread over many samples: at the same total number of repeats,
concentrating them on fewer samples costs more. In the 45M partial-repetition study
(\S\ref{sec:surface}), the allocations $(f,R_{\rm sub})=(50\%,2)$ and $(33.3\%,4)$ match model size,
budget, distinct targets and total extra exposures; they differ only in whether each repeated target is
seen once more on a larger subset or three times more on a smaller one. Concentrating the repeats raises
loss by 0.004~bpb (Appendix Figure~\ref{fig:replacement}d).}}

\begin{figure}[t]\centering
\includegraphics[width=\linewidth]{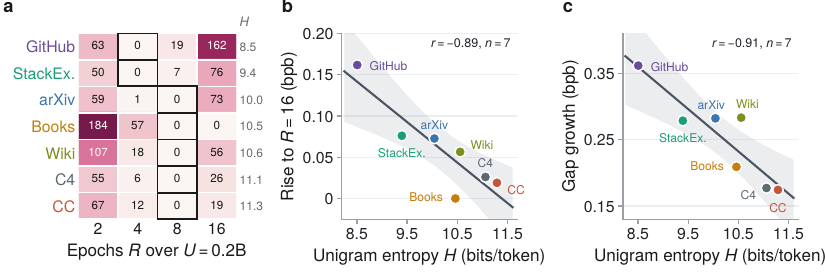}
\caption{\textbf{\revise{Lower-entropy sources exhibit steeper post-minimum degradation.}}
\revise{Seven SlimPajama sources trained separately (59M, $U=0.2$B), each evaluated on its own held-out set.} \arxnew{(a)~Loss above each source's minimum, in $10^{-3}$~bpb; rows are ordered by unigram entropy $H$ (bits/token, right), and boxes mark the loss-minimizing $R$ ($R=1$ omitted).} \revise{(b)~Rise from minimum to $R=16$ and (c)~unseen-minus-seen gap growth ($R=1\to16$) vs.\ unigram entropy; \arxnew{lines are least-squares fits with 95\% confidence bands}.}}
\label{fig:entropy}
\end{figure}

\arxnew{\paragraph{Source.}\label{sec:source}
\arxlatest{Lower-entropy sources tolerate less repetition: they degrade more past their loss minimum, so the
same epoch count costs more on code than on prose. We train a 59M model separately on each of the seven
sources at $U=0.2$B. Every source improves through four or eight epochs, but lower unigram entropy
predicts a larger rise in loss by 16 epochs and faster growth of the unseen-minus-seen gap ($r=-0.89$ and
$-0.91$; Figure~\ref{fig:entropy}); a tokenizer-independent compression ratio gives the same ordering.
GitHub, the lowest-entropy source, ends 0.16~bpb above its minimum, whereas Books does not rise. GitHub \arxlatest{also
pays the most for full-corpus repetition in the 45M study and reaches its minimum earlier than CommonCrawl at
$U=2.5N$} (\S\ref{sec:fixed_pool}).}}

\subsection{\texorpdfstring{\arxnew{Re-tokenizing repeated data}}{Re-tokenizing repeated data}}
\label{sec:bonus}\label{sec:bpe_intervention}
\begin{figure}[t]\centering
\includegraphics[width=\linewidth]{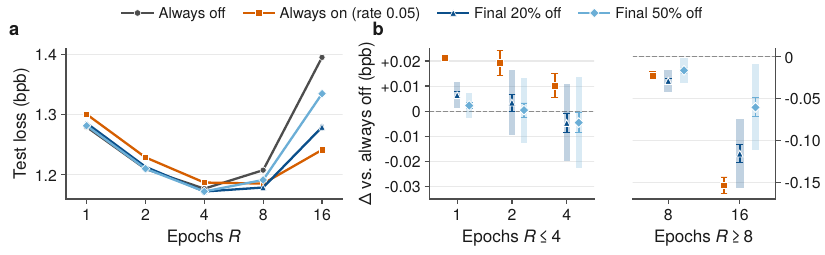}
\caption{\textbf{\arxlatest{Re-tokenizing repeats raises loss at
$R\le4$ and lowers it at $R\ge8$.}} (a)~Final test loss: means and replicates. (b)~Paired
difference from always-off with pointwise 95\% $t$ intervals; light bars show
simultaneous Bonferroni intervals \arxlatest{for the policy contrasts}. \revise{$R\le4$
and $R\ge8$ use separate linear scales.}}
\label{fig:bpe_policies}
\end{figure}

A repeated epoch shows the model the same bytes, but not necessarily the same
tokens. Stochastic byte-pair encoding (BPE-dropout;
\citealp{provilkov2020bpe}) re-segments each occurrence
while preserving bytes, so a repeat can present new tokens without new data. \arxlatest{Training 45M models on CommonCrawl for $R\in\{1,2,4,8,16\}$ epochs, we compare four schedules:
always-off, always-on (\arxlatest{dropout rate} 0.05), and on until the final 20\% or 50\% of training
(Figure~\ref{fig:bpe_policies}). All schedules see the same bytes in the same order and are evaluated
with the deterministic tokenization (Appendix~\ref{app:bpe_methods}).}

\arxlatest{Re-tokenization helps only when reuse is heavy. Relative to always-off, always-on raises loss by
0.021 and 0.010~bpb at $R=1$ and 4 but lowers it by 0.022 and 0.154~bpb at $R=8$ and 16. From 4 to 16
epochs it cuts the rise in loss from 0.218 to 0.054~bpb, attenuating the cost of heavy reuse without
removing it. Switching it off for the final 20\% of training improves on always-on by 0.015~bpb at
$R\le4$ but is worse by 0.038~bpb at $R=16$, so no single schedule is best at every horizon.}

\section{\texorpdfstring{\arxlatest{Discussion}}{Discussion}}
\label{sec:discussion}
\arxnew{\paragraph{Implications for pretraining on finite data.}
\arxlatest{The four takeaways in \S\ref{sec:intro} rest on the following results.}}
\begin{itemize}
\item \arxlatest{\emph{Epochs and model size.} On the grid, model size enters the cost of repetition only through
$z=(R-1)N_{\rm total}/U$. With $U$ fixed, $z$ grows with $N$, and the loss minimum falls from about 15
epochs at 127M to 4 at 2B (\S\ref{sec:rpeak}); with $U/N$ fixed, it stays near four epochs on
CommonCrawl, though on GitHub it drifts down (\S\ref{sec:fixed_pool}). To transfer an epoch count to a
larger model, scale the unique data with it.}
\item \arxlatest{\emph{Budget and tolerance.} Repeated tokens fall to half the value of fresh ones after
$R_c\approx2.7\,(D/N)^{0.24}$ epochs, about four to five up to a TPP of 20 and ten at 300. $R_c$ rises
because the cost of a repeat falls with unique data per parameter faster than the fresh-data gain it
forgoes (\S\ref{sec:two_readings}).}
\item \arxlatest{\emph{Allocation at fixed unique data.} The fitted frontier holds TPP slightly below the
fresh-data optimum, so $N$ and $R$ each grow roughly as $\sqrt{C}$; loss is lowest near $R_c$, and
additional compute beyond that point does not help. Ignoring the cost of repetition loses little near the
optimum (\S\ref{sec:frontier}).}
\item \arxlatest{\emph{How repeats are laid out.} At identical counts, spacing repeats across epochs and
spreading them over many samples lowers loss, and lower-entropy sources tolerate fewer epochs
(\S\ref{sec:schedule}); re-tokenizing repeats helps only under heavy repetition
(\S\ref{sec:bonus}). Compare such choices at separately annealed horizons, since intermediate
checkpoints can misrank them (Appendix~\ref{app:allocation}).}
\end{itemize}

\arxlatest{\paragraph{Limitations and future work.}
Our models are far smaller than frontier systems, and all runs \arxlatest{use recipes} without
weight decay or dropout; since weight decay lowers the cost of repetition \citep{lovelace2026prescriptive}, the constants
of Eqs.~\eqref{eq:tpp_grid} and~\eqref{eq:rc} may shift under other regularization. We measure held-out
loss rather than downstream capabilities. Natural next steps are larger models with iso-FLOP runs that
locate the loss minimum directly, more complex data mixtures in which sources are repeated at different rates,
and downstream evaluation.}

\section{\texorpdfstring{\arxlatest{Conclusion}}{Conclusion}}
\label{sec:conclusion}
\arxlatest{We asked what a repeated token is worth. Measured against fresh data at equal compute, its cost
follows a single combination of counts, $z=(R-1)N_{\rm total}/U$. Together with the one-epoch curve,
this variable \arxlatest{fixes the geometry of finite-data scaling: it accounts for the rise of} the critical epoch count at which repeats fall to half the value of fresh tokens,
the split of a fixed pool of unique data between model size and epochs, and the direction of the size
trend under each experimental design. Beyond counts, the value of a repeat depends on how the repeats are
ordered, allocated, sourced and tokenized. Scaling studies of finite-data pretraining should therefore
report the baseline, the quantity held fixed and these choices alongside the counts.}

%% file: related-work-arxiv.tex
\paragraph{Reuse under data constraints.}
Compute-optimal scaling allocates compute between parameters and fresh tokens
\citep{kaplan2020scaling,hoffmann2022training}. Data-constrained scaling
assigns repeated tokens a diminishing effective value
\citep{muennighoff2023scaling}, and multi-epoch degradation has been reported
to worsen with model size \citep{xue2023repeat}. \citet{lovelace2026prescriptive}
add a penalty in $(R-1)N/U$ (with embeddings in $N$) and derive compute-optimal
allocations that turn toward fewer epochs at large compute, and
\citet{qin2026bridging} define the token effectiveness $\eta$ of derived tokens and model it
across model size, fresh tokens per parameter and repetition. \citet{yan2026larger} analyze reuse
efficiency in linear regression. \revise{At a fixed compute budget,
\citet{muennighoff2023scaling} find up to about four epochs nearly as valuable as
fresh data;} \arxnew{we find that this number grows with the training budget per parameter
(\S\ref{sec:two_readings}). We read the value and cost of repetition directly off measured runs,
against two baselines in the same models: one epoch on the same unique data, and fresh data at equal
compute.}

\paragraph{Partial repetition and exposure structure.}
Repeating a data subset can degrade models
\citep{hernandez2022scaling}, and deduplication improves them
\citep{lee2022deduplicating}. The closest fixed-budget \revise{study}
\citep{chudnovsky2026fr} replays 10\% of tokens up to 20,000 times and studies
the repeat count that \emph{maximizes} loss. \citet{kazdan2026duplication}
relate semantic redundancy to replay in finite-corpus pretraining. \arxlatest{Our partial-repetition studies}
cover lower repeat counts across sources and fractions and \arxlatest{add a control that concentrates
repeats at matched counts}. Studies of memorization under repetition
\citep{carlini2023quantifying,tirumala2022memorization} are complementary: we
measure held-out loss rather than extraction.

\paragraph{Recipes and mixtures.}
Regularization, ensembling and data augmentation change the attainable
loss under a fixed corpus
\citep{kim2025infinite,xu2026dataconstrained,chen2026augmentations}.
Sparsity \citep{wu2026sparse} and memorization-guided epoch and replay
schedules \citep{zuo2026train} also matter. We hold one recipe fixed across
allocations.
Mixture studies address source composition and per-source repetition directly
\citep{liu2026infolaw,sedova2026mixture,zhou2026mismatch}. Data quality and filtering laws also characterize the information value of distinct corpora \citep{goyal2024scaling}, while compression-based metrics link data redundancy to scaling exponents \citep{pandey2024gzip}. \arxlatest{We train sources separately and do not optimize mixtures.}

\paragraph{Stochastic tokenization.}
Subword regularization
\citep{kudo2018subword} and BPE-dropout \citep{provilkov2020bpe} were
introduced for translation. BPE-dropout is used in \revise{large-scale
pretraining to reduce overfitting to frequent patterns
\citep{ernie2026ernie5}} and as a baseline for stochastic token splitting
\citep{sims2026stochastok}. It also appears in masked-LM studies
\citep{visser2026subword} and mitigates LLM sensitivity to typographical and
format variation \citep{chai2024tokenization}. We use BPE-dropout and late
shutoff as existing tools and measure their effect on repeated data across
horizons\arxnew{; because re-segmentation splits the same bytes into more tokens, we also report the
extra tokens and compute it costs.}

%% file: appendix-arxiv.tex
\section{Architectures, Training Recipes and Evaluation Protocols}
\label{app:new_hparams}\label{app:protocols}
\arxlatest{Table~\ref{tab:families} summarizes the experiments and the quantities each holds fixed. This section
gives the architectures, training recipes and evaluation protocol they share.}
\input{tables/experiment-families-arxiv.tex}

\begin{table}[!htbp]\centering\small
\caption{\arxlatest{Architectures of the paired 45M and 185M models. $F$ counts analytical training
FLOPs per target.}}
\label{tab:architectures}
\begin{tabular}{lrr}\toprule
Field &45M model&185M model\\\midrule
Exact non-embedding parameters&45,084,900&184,623,360\\
Tied embedding parameters&32,164,480&51,463,168\\
Exact total parameters&77,249,380&236,086,528\\
Rounded total size&77.2M&236.1M\\
Width / depth&640 / 10&1024 / 16\\
Width/depth ratio&64&64\\
Query / KV heads&10 / 5&16 / 8\\
Head dimension&64&64\\
MLP width / activation&2560 / ReLU$^2$&4096 / ReLU$^2$\\
Vocabulary / context length&50,257 / 1024&50,257 / 1024\\
$F$ per target&502,644,480&1,516,738,560\\\bottomrule
\end{tabular}\end{table}

The paired models use grouped-query attention, rotary positions (base 10,000),
RMS normalization, tied embeddings initialized at standard deviation 0.005,
query/key gain 1.5 and logit softcap 30. Hidden matrices use Muon with five
Newton--Schulz steps; scalar/vector and tied-embedding parameters use Adam,
$(\beta_1,\beta_2)=(0.9,0.95)$ and $\epsilon=10^{-8}$. Weight decay, dropout and
gradient clipping are zero. \arxlatest{Parameters and optimizer states are kept in FP32 and activations in
BF16; models are trained on eight H800 GPUs with PyTorch 2.9.} Table~\ref{tab:paired-config}
lists the per-study batch, accumulation and update settings.

\input{tables/paired-configurations-arxiv.tex}

Paired-run momentum increases from 0.85 to 0.95 over $\lceil 0.1K\rceil$ updates,
and LR decays linearly over the final $\lceil 0.2K\rceil$, with $K=D/B$.
In contrast, the primary broad sweeps use $(0.04,0.04,0.05)$ and ramp momentum
over 500 updates, with global batches in Table~\ref{tab:ladder}.

\paragraph{\arxlatest{Learning-rate calibration.}}
The 12 source-study calibration runs compare $(0.01,0.01,0.05)$ at
$B=131{,}072$ with $(0.02,0.02,0.05)$ at $B=131{,}072$ and 262,144 on
CommonCrawl/GitHub, for fresh data and for \revnew{$(50\%,8)$}\arxlatest{; the lower rate lies within
0.002~bpb of the best and is retained}. Sixteen scale calibration runs compare the two LR tuples
on both sizes and sources, for fresh data and 12 full-corpus epochs\arxlatest{; $(0.02,0.02,0.05)$ is best
(mean regret 0 versus 0.0119~bpb)}. Calibration runs are excluded from all reported comparisons.

\paragraph{\arxlatest{Matched budgets in the two-size study.}}
\arxlatest{The 45M and 185M models of Appendix~\ref{app:paired_scale} match $D/N$ and
$U/N$ to within 0.03\% in non-embedding parameters; counted with embeddings, $D/N$ is 2.6 and 3.5.
Because the batch grows fourfold with size, the two models take \arxlatest{1,536 and 1,572 updates at matched $D/N$
($384R$ and $393R$ at matched $U/N$)}.}

\begin{table}[!htbp]\centering\small
\caption{\arxlatest{\textbf{The proxy family.} $N$ is the $12Ld^2$ size label and $N_{\rm count}$ the exact
non-embedding count in millions; batch is in target tokens per update. All models have head dimension 64
and MLP width $4d$. The mixed-corpus ladder of the fixed-budget sweep (Figure~\ref{fig:ds}d) shares the
31M and 59M rows; its 252M model is also used in the 252M partial-repetition panel.}}
\label{tab:ladder}
\begin{tabular}{@{}rrrrrrr@{}}
\toprule
$d$ & $L$ & $d/L$ & heads (KV) & $N$ & $N_{\rm count}$ (M) & batch (tok) \\
\midrule
\multicolumn{7}{l}{Main CommonCrawl model-size sweep} \\
384  & 8  & 48.00 & 6 (3)   & 14M  & 12.990   & 262{,}144 \\
512  & 10 & 51.20 & 8 (4)   & 31M  & 28.859   & 524{,}288 \\
640  & 12 & 53.33 & 10 (5)  & 59M  & 54.102   & 524{,}288 \\
768  & 18 & 42.67 & 12 (6)  & 127M & 116.848  & 524{,}288 \\
1024 & 20 & 51.20 & 16 (8)  & 252M & 230.779  & 524{,}288 \\
1280 & 26 & 49.23 & 20 (10) & 511M & 468.733  & 1{,}048{,}576 \\
2048 & 20 & 102.40 & 32 (16) & 1.0B & 922.932 & 1{,}048{,}576 \\
2560 & 26 & 98.46 & 40 (20) & 2.0B & 1874.630 & 1{,}048{,}576 \\
\midrule
\multicolumn{7}{l}{Mixed-corpus ladder} \\
768  & 16 & 48.00 & 12 (4)  & 113M & 100.719  & 524{,}288 \\
1024 & 20 & 51.20 & 16 (4)  & 252M & 220.293  & 1{,}048{,}576 \\
1280 & 24 & 53.33 & 20 (5)  & 472M & 413.016  & 1{,}048{,}576 \\
\bottomrule
\end{tabular}
\end{table}

\paragraph{\arxlatest{Size labels and metrics.}}
\arxlatest{The proxy $N=12Ld^2$ differs from exact non-embedding counts and does not fix the architecture: at
the 252M label, the four-KV partial-repetition runs count 220.3M non-embedding parameters and the
eight-KV ladder 230.8M. The sweeps report a byte-normalized metric; the paired studies
report bpb over ordinary targets on their own holdouts.}

\paragraph{Single-size optimizer comparison.}
\label{app:optimizer_sensitivity}
\arxlatest{Twelve 59M-proxy runs, six with Muon and six with AdamW \citep{jordan2024muon,loshchilov2019adamw}, share one architecture,}
mixed corpus, $D=3.2$B and batch 524,288; each repeat count uses $U=D/R$.
Both ladders worsen relative to their own one-epoch baseline (Table~\ref{tab:optimizer_sensitivity}). At $R=32$,
\arxlatest{Muon and AdamW reach 1.868 and 1.761~bpb, although Muon's one-epoch loss is lower (1.113 versus 1.153)}\arxlatest{: the high-reuse degradation persists under
both optimizers}.
\begin{table}[!htbp]\centering\small
\caption{\textbf{Optimizer sensitivity at one model size and token budget.}
Absolute \arxlatest{mixed-corpus} losses $L$ are in bpb; $\Delta L=L(R)-L(1)$
is computed relative to each optimizer's own single-epoch endpoint, in
$10^{-3}$ bpb.}
\label{tab:optimizer_sensitivity}
\begin{tabular}{@{}rrrrr@{}}
\toprule
$R$ & $L_{\rm Muon}$ & $L_{\rm AdamW}$ & $\Delta L_{\rm Muon}$ & $\Delta L_{\rm AdamW}$ \\
\midrule
1 & 1.112899 & 1.152695 & 0.000 & 0.000 \\
2 & 1.114845 & 1.153794 & 1.945 & 1.100 \\
4 & 1.124383 & 1.157711 & 11.484 & 5.016 \\
8 & 1.160769 & 1.189120 & 47.869 & 36.426 \\
16 & 1.279985 & 1.283226 & 167.086 & 130.531 \\
32 & 1.867810 & 1.760766 & 754.911 & 608.072 \\
\bottomrule
\end{tabular}
\end{table}
\paragraph{\arxlatest{Evaluation sets.}}
\arxlatest{Each source has a development and a test split of 1,024 non-overlapping windows of 1,025 tokens
(1,048,576 targets), drawn with a fixed seed; any training window that matches a held-out window in full
is removed. The scale studies use the same CommonCrawl and GitHub splits. We report bits per byte over
ordinary targets; end-of-text tokens count toward the training budget $D$ but not toward the metric.}

\FloatBarrier
\section{\texorpdfstring{\arxlatest{Paired Estimates and Statistical Tests}}{Paired Estimates and Statistical Tests}}
\label{app:paired}\label{app:adjacent}
\arxlatest{A replicate is one pair of data-order and initialization seeds; a paired contrast differences two
runs that share it, so seed effects common to both cancel. The partial-repetition and scale studies use
three replicates ($t_{0.975,2}=4.303$); the concentration contrast uses six ($t_{0.975,5}=2.571$) and an
exact sign-flip test over all 64 sign patterns.} Development data alone fixed recipes and repeat
candidates; no test outcome selected a checkpoint, seed, repeat count or fit. \arxlatest{Paired studies report
95\% $t$ intervals over replicates. Single-run sweeps use a resolution threshold of 0.002~bpb, below which
differences are treated as unresolved. The grid, the fixed-$U$ sweeps, the $U=2.5N$ study and the source
sweeps have one replicate per configuration, except for two extra seeds at 78M, $D/N=100$, $R\le2$
(Appendix~\ref{app:grid}).}

\input{tables/paired-summary-arxiv.tex}

Full-corpus candidates are selected on development means before test access:
45M/CC four, 45M/GitHub two, and both 185M sources four epochs. The
20 adjacent contrasts share simultaneous 95\% Bonferroni paired-$t$ intervals
(critical value 19.962, df 2)\arxlatest{; with three replicates, no candidate is bracketed by significantly
worse neighbors}.

\paragraph{\arxlatest{Decision rules for the two-size study.}}
The primary statistic is the two-source mean of the
larger-minus-smaller excess loss under partial repetition at \revnew{$(f,R_{\rm sub})=(50\%,8)$}; the
\revnew{$(50\%,2)$} contrast is secondary. Practical equivalence requires the whole
interval inside $\pm0.005$ bpb. For full-corpus repetition, the three-replicate development mean selects a candidate from $\{1,2,4,6,8,12\}$, with ties
favoring fewer epochs, before any test access; a local bracket requires both
neighboring losses to be higher under the simultaneous intervals.

\FloatBarrier
\section{\texorpdfstring{\arxnew{The $D/N\times R$ Grid}}{The D/N x R Grid}}
\label{app:grid}\label{app:qin}
\paragraph{Sizes and recipe.}
The grid uses the architecture family of the scale studies (Appendix~\ref{app:scale_new}) at five
exact non-embedding sizes, 23,087,168, 45,084,900, 77,898,384, 184,623,360 and 360,563,600
parameters, with tied embeddings over the 50,257-token vocabulary\arxnew{, which $N_{\rm total}$ counts
once}. Every size is trained at
$D/N\in\{5,10,20,50,100,200,300\}$ and $R\in\{1,2,4,8,16\}$ over $U=D/R$ CommonCrawl targets. Runs of
one size and $D$ share recipe, step count and schedule, so the repeated run and its fresh baseline
differ only in the data. \arxnew{Training uses Muon without learning-rate warmup or weight decay,
with a constant learning rate followed by a final warmdown.} Each epoch is an independent shuffle of
the same $U$ targets.
\arxnew{Each size also has five shorter one-epoch runs, $D/N\in\{0.3125,0.625,1.25,2.5,3.125\}$, the
unique budgets $D/R$ of the repeated runs below $D/N=5$, with the learning rate from the same rule.}

\paragraph{Corpus extension.}
Nine runs need more unique data than the 25.8B-token CommonCrawl pool. They use an extended pool that
keeps the original as an exact prefix and shares its test set, so runs with smaller $U$ see the same
data under either pool.

\begin{figure}[!htbp]\centering
\includegraphics[width=\linewidth]{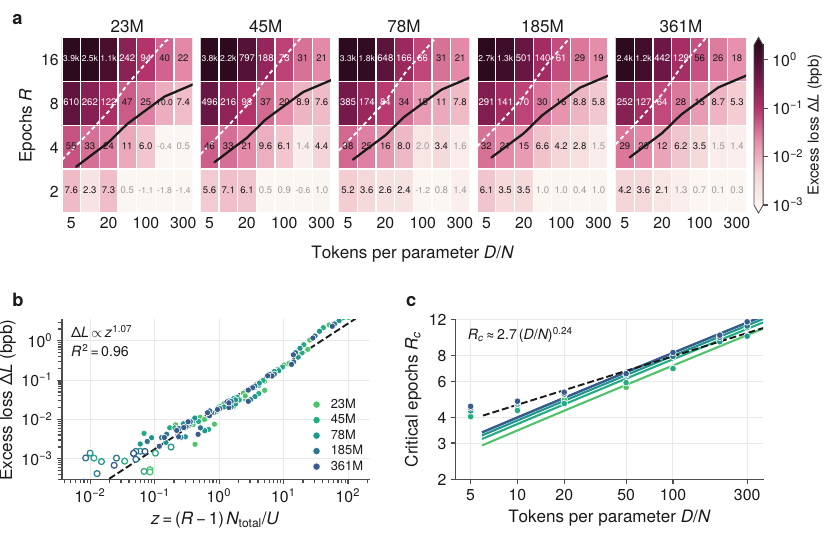}
\caption{\arxnew{\textbf{\arxlatest{The grid at every size, with values (cf.\ Figure~\ref{fig:rc}).}} (a)~Excess loss of $R$ epochs over $U=D/R$
against one epoch over $D$ fresh tokens at the same $N$ and $D$, in units of $10^{-3}$~bpb. Black
line, \arxlatest{$R_c$ from the fresh-data fit plus Eq.~\eqref{eq:tpp_grid} (solid lines in panel c)}; dashed, the 0.1-bpb level set of
Eq.~\eqref{eq:tpp_grid}. (b)~All 140 repeated configurations against $z$; open markers fall below
the 0.002-bpb resolution threshold and are not fitted. (c)~Critical epoch count $R_c$, at which
$\eta=\tfrac12$ (Eq.~\ref{eq:ceg}): measured crossings, interpolated in $\log R$ (points); the
fresh-data fit plus Eq.~\eqref{eq:tpp_grid} (lines); and Eq.~\eqref{eq:rc} fitted to the points
(dashed).}}
\label{fig:map}
\end{figure}

\paragraph{Resolution.}
Three seeds of the 78M, $D/N=100$ configuration give two-epoch excess losses of 0.0003--0.003~bpb, so
differences below 0.002~bpb are treated as unresolved. Runs below this threshold are shown hollow in
Figure~\ref{fig:map}b and excluded from the fit of Eq.~\eqref{eq:tpp_grid}\arxnew{, which uses the
remaining 113 of the 140 repeated runs. Two-epoch excess losses lie close to this threshold, so the
two-epoch values of $\eta$ in Figure~\ref{fig:fte} are the least precise}.

\begin{figure}[!htbp]\centering
\includegraphics[width=0.92\linewidth]{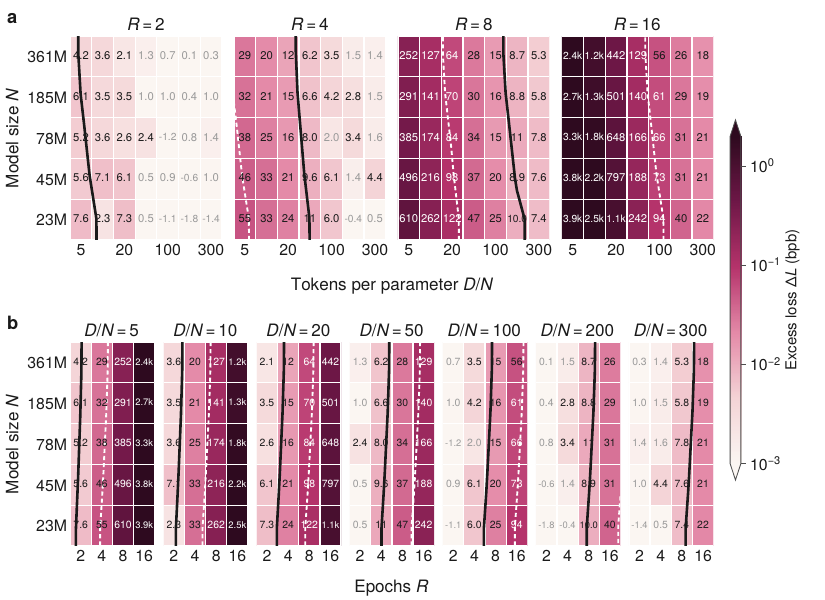}
\caption{\arxnew{\textbf{The grid re-sliced by epochs and by budget.} Excess loss of $R$ epochs over
one epoch at the same $N$ and $D$, in units of $10^{-3}$~bpb, (a)~at fixed $R$ and (b)~at fixed $D/N$;
Figure~\ref{fig:map}a gives the slices at fixed model size. Solid and dashed lines are the 0.01- and 0.1-bpb contours of
Eq.~\eqref{eq:tpp_grid}; gray numbers fall below the 0.002-bpb resolution
threshold.}}
\label{fig:grid_slices}
\end{figure}

\paragraph{Fits.}
$A$ and $p$ are fitted by least squares in $\log\Delta L$ over the runs above the threshold; the size
term multiplies Eq.~\eqref{eq:tpp_grid} by $N_{\rm total}^{-g}$. $\eta$ inverts each size's one-epoch
curve, interpolated linearly in $\log D$ and $\log L$; runs whose loss is above the \arxnew{$D/N=0.3125$}
one-epoch loss or whose $D$ exceeds the one-epoch range are skipped, and budgets with fewer than
three sizes are not plotted. \arxlatest{Figure~\ref{fig:map} gives the values behind Figure~\ref{fig:rc}, every repeated configuration
against $z$ and the fit of Eq.~\eqref{eq:rc} to the measured crossings; unlike Figure~\ref{fig:rc}, its
panel (a) draws $R_c$ from the fresh-data fit plus Eq.~\eqref{eq:tpp_grid}.} Figure~\ref{fig:grid_slices} re-slices
the grid of Figure~\ref{fig:map}a by $R$ and by $D/N$.

\begin{figure}[!htbp]\centering
\includegraphics[width=\linewidth]{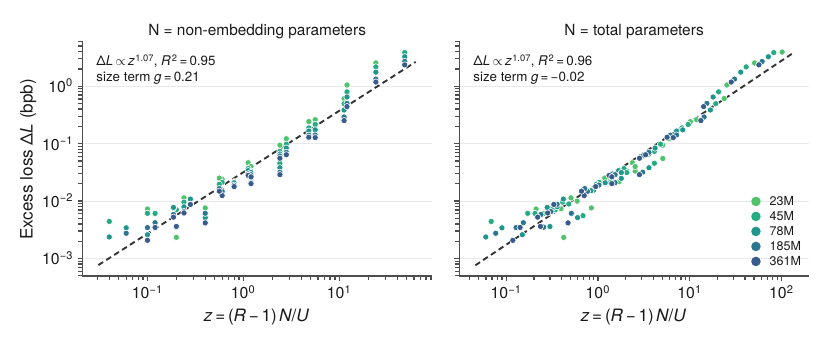}
\caption{\textbf{Counting embeddings removes the size term.} Grid excess loss (bpb) against
$z=(R-1)N/U$ with $N$ the non-embedding (left) or total (right) parameter count.}
\label{fig:grid_collapse}
\end{figure}

\begin{figure}[!htbp]\centering
\includegraphics[width=\linewidth]{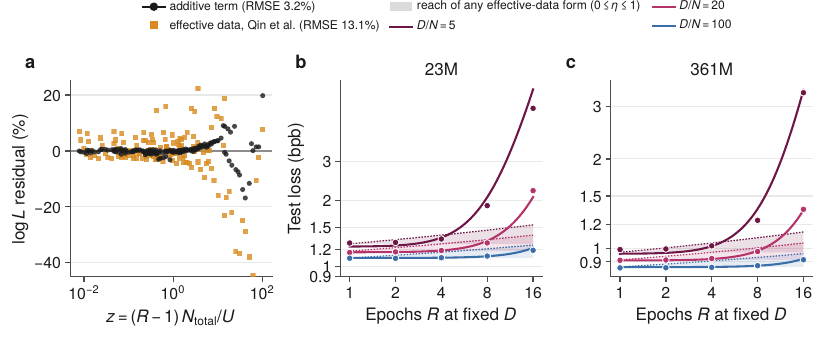}
\caption{\textbf{An additive repetition term fits the grid; an effective-data form cannot reach the
high-$z$ runs.} (a)~Residuals in $\log L$ of the joint additive fit (black) and of the joint
effective-data fit of \citet{qin2026bridging} (orange) against $z$. (b,c)~Test loss at fixed $D$
against $R$ for three budgets at 23M and 361M: measured runs (points) and the additive fit (lines).
Shading spans every effective-data law with $0\le\eta\le1$ on the one-epoch backbone; its lower edge
is $\eta=1$ (fresh data) and its dotted upper edge $\eta=0$ (one epoch over $U$), extrapolated where
$U<5N$.}
\label{fig:grid_laws}
\end{figure}

\paragraph{Joint fits.}
Adding the repetition term to a Chinchilla backbone\arxnew{, in the additive form of
\citet{lovelace2026prescriptive}}, $L=E+A'N_{\rm total}^{-\alpha}+BD^{-\beta}+\arxround{G}z^{p'}$,
fits all 175 runs to \arxnew{3.2}\% RMS in $\log L$ (Figure~\ref{fig:grid_laws}). The exponent of this joint fit
($p'=\arxnew{1.25}$) differs from that of the excess fit ($p=\arxnew{1.07}$). The main text uses the excess fit,
which needs no backbone because the same-budget fresh baseline cancels it. The effective-data form of \citet{qin2026bridging},
$L=E+A'N_{\rm total}^{-\alpha}+B\,[U+\eta(D-U)]^{-\beta}$, fits \arxnew{13.1}\% RMS. At fixed $D$ it can raise
loss only by lowering effective data, which is bounded below by $U$; to reproduce the steep rise of
loss at small budgets and high $R$, the fit raises $\beta$ to \arxnew{1.04}, against 0.19 from the
one-epoch runs alone, and then misfits those runs too: at 23M and $D/N=5$ it predicts \arxnew{1.79}~bpb
for the fresh run, which measures \arxnew{1.28}.

The shortfall does not depend on the fitted parameters. Any law of the form
$L=L_{\rm fresh}(U+\eta(D-U))$ with $0\le\eta\le1$ lies between $L_{\rm fresh}(D)$ and $L_{\rm fresh}(U)$,
the band in Figure~\ref{fig:grid_laws}b,c. With $L_{\rm fresh}$ fitted to the 35 one-epoch runs
($E=0.583$, $\alpha=0.305$, $\beta=0.188$; RMSE 0.86\%), 30 of the 140
repeated runs lie above the upper edge: every run with $z\ge5.7$ except one ($z=5.9$); letting
$\eta$ vary freely within $[0,1]$ then leaves an RMSE of \arxnew{28.0}\% on the repeated runs. These runs
all have $U<5N$, below the seven grid budgets of the fit. \arxnew{The short one-epoch runs measure
$L_{\rm fresh}(U)$ there directly: of the 80 repeated runs with a one-epoch run at the same $U$,
21 lie above it, by up to 3.21~bpb, namely all 15 with $z>11.3$ (16 epochs, $D/N\le20$), four of the
five 8-epoch runs at $D/N=5$ ($z\approx11.2$) and the 16-epoch runs of the two smallest models at
$D/N=50$ ($z\approx4.8$, by at most 0.02~bpb); every other repeated run is below it.} The additive term has no such bound and follows the rise.

\paragraph{Inverting a joint fit.}
Effective data can also be read off a fitted law by solving it for the $D_{\rm eff}$ that reproduces
each repeated run. A loss residual $\delta L$ then becomes a change in $D_{\rm eff}$ of order
$\delta L\,D^{1+\beta}/(\beta B)$, so residuals of a few $10^{-3}$~bpb move the inferred efficiency by large
amounts at large budgets. On the grid, efficiencies inverted in this way show no consistent trend with
TPP, whereas the values read off measured one-epoch curves rise with it
(Figure~\ref{fig:fte}). \citet{qin2026bridging} report efficiency falling with tokens per parameter\arxlatest{; their TPP counts fresh
tokens only, but at fixed $R$ it is proportional to ours, so the two directions do conflict}. Besides the
method, their regime differs, with weight decay tuned per run up to 1.6, against none here.

\arxnew{\paragraph{Critical epoch count and frontier.}
\arxnew{Each crossing interpolates the measured $\eta$ in $\log R$ between neighboring epoch counts.
The 35 crossings, one for each size and budget, follow Eq.~\eqref{eq:rc} to 8\% RMS. $R_c$ is
4.0--5.3 up to a TPP of 20, 6.9--8.3 at 100 and 10.0--11.8 at 300, so a single power law understates
the flattening at small budgets.} Adding a size term $N^{0.04}$ to Eq.~\eqref{eq:rc}
barely changes the fit. \arxnew{The crossings use the short one-epoch runs, as in
Figure~\ref{fig:fte}; at a TPP of 5 and 10 they fall between $R=4$ ($\eta\approx0.6$) and $R=8$
($\eta\le0.1$), so they mainly reflect $\eta$ at four epochs.} The one-epoch fit plus
Eq.~\eqref{eq:tpp_grid} gives a slope of 0.31; \arxnew{its crossings lie 27--34\% below the measured
ones at a TPP of 5 and} within 15\% at 300. The one-epoch backbone used
for the frontier fits the 35 one-epoch runs with an RMSE of 0.0024~bpb. Pooled over
$U\in\{0.5,1,1.5\}$B \arxlatest{and the full interior frontier}, $R_{\rm opt}\propto C^{0.45}U^{-0.83}$, \arxlatest{flatter than at each $U$ because it includes the stretch beyond the loss minimum}. \arxlatest{Up to the lowest predicted loss, $D/N$ drifts from 22 to 16 at
$U=0.5$B, \arxlatest{from 23 to 19 at 1B} and from 24 to 21 at 1.5B. That minimum lies at 164M and 5.3 epochs ($C=2.6\times10^{18}$)
at $U=0.5$B, \arxlatest{at 297M and 5.6 epochs ($C=9.9\times10^{18}$) at 1B} and at 361M and 6.1 epochs ($C=2.0\times10^{19}$) at 1.5B. Past it, the frontier at
$U=0.5$B continues to 7 epochs at higher loss before $N_{\rm opt}$ reaches the largest grid model.} The fixed-$U$ size exponent of 0.70 comes from six approximate IsoFLOP minima at
$U=1$B.}

\begin{figure}[!htbp]\centering
\includegraphics[width=\linewidth]{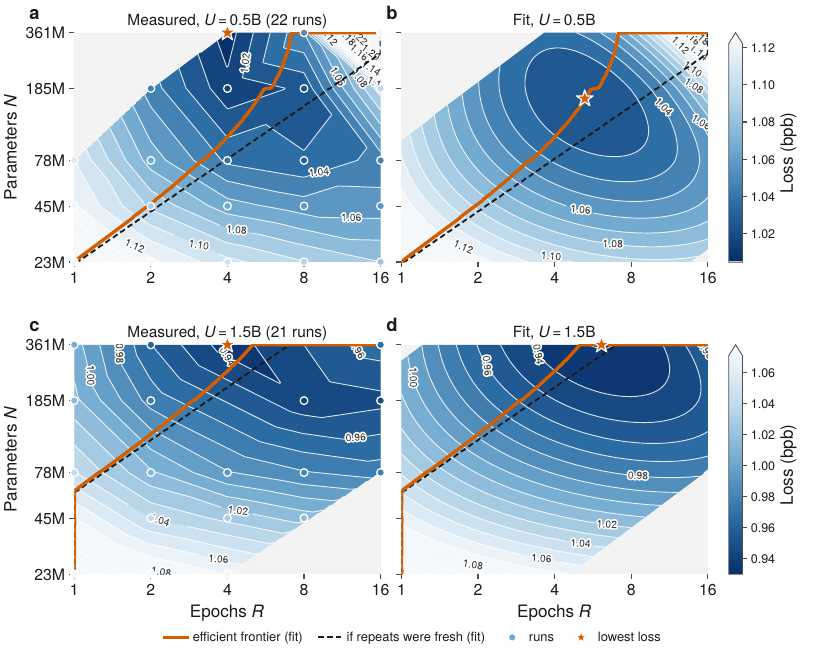}
\caption{\textbf{Iso-loss contours at fixed unique data.} Test loss over epochs and model size at
$U=0.5$B (a,b) and 1.5B (c,d). (a,c)~Linear interpolation, in $\log N$ and $\log R$, of grid runs whose
$U$ lies within a factor 1.41 of the target, each shifted to the target $U$ by the fit. (b,d)~The fit
used for Figure~\ref{fig:frontier}: one-epoch backbone plus Eq.~\eqref{eq:tpp_grid}, shown where
$5\le D/N\le300$. Contours every 0.01~bpb. Solid orange, the efficient frontier; dashed, the frontier
if repeats were fresh; stars, the lowest measured or fitted loss. Runs differ from the fit by
\arxnew{0.011} and \arxnew{0.002}~bpb on average at the two $U$. The lowest measured loss is at 361M and
$R=4$ at both $U$; the fit places its minimum at $R\approx5$--6.}
\label{fig:gridisoloss}
\end{figure}

\section{Fixed-Budget Sweeps and the 59M Example}
\label{app:fixed_budget}
\begin{figure}[!htbp]\centering
\includegraphics[width=\linewidth]{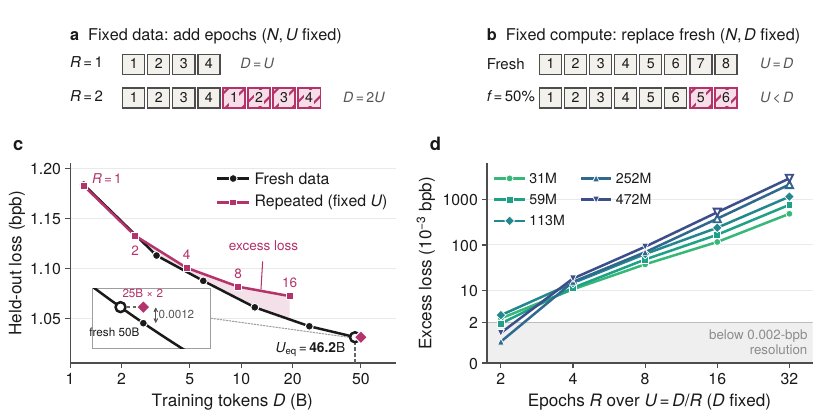}
\caption{\revise{\textbf{At the 59M proxy size, a second epoch retains most of a fresh epoch's value (c); at a
fixed token budget, repetition raises loss at every model size (d).}
(a)~Fixed-corpus repetition keeps model size $N$ and unique tokens $U$ fixed, so extra epochs add
training tokens $D=RU$ and compute. (b)~\arxlatest{Partial repetition} keeps $N$ and $D$ fixed; a
fraction $f$ of training tokens comes from a subset seen $R_{\rm sub}$ times, so
$U=D(1-f+f/R_{\rm sub})$. (c)~59M proxy size: one-epoch fresh-data endpoints (circles) and a fixed
$U=1.2$B ladder labeled by $R$ (squares); shading marks the excess loss between them. The diamond
is 25B repeated twice; dashed guides map it to $U_{\rm eq}=46.2$B on the reference
$L=E+AD^{-\alpha}$ fitted to the 6--50B endpoints (hollow circle), which gives $\eta$
(Eq.~\ref{eq:ceg}); the inset resolves the 50B endpoints. (d)~Excess loss of $R$ epochs over
$U=D/R$ versus one epoch at the same model and $D$ (3.2B; 4B for 252M/472M at $R\le8$; open
markers \arxlatest{give the two largest ladders at $D=3.2$B, $R=16$ and 32}); the axis is linear below $0.002\text{ bpb}$ (gray
band). Five proxy sizes from 31M to 472M.}}
\label{fig:study_design}\label{fig:ds}
\end{figure}

\paragraph{\arxnew{A long run at 59M.}}
\revise{At the 59M proxy size}, two epochs over 25B unique tokens reach
1.0308 bpb, against 1.0296 for one epoch over 50B fresh tokens
(Figure~\ref{fig:study_design}c)\revise{, a gap below the $0.002\text{ bpb}$ threshold}.
\mbox{Inverting} \revise{a power law fitted to four fresh-data endpoints} assigns the repeated run 46.2B
fresh-equivalent tokens, so the second epoch contributes about 21.2B
($\eta=0.85$; $U_{\rm eq}/D=0.92$ over the whole run). \arxlatest{Because the fresh-data curve is flat at this budget, perturbing the repeated endpoint by
$\pm0.002$~bpb moves $\eta$ over 0.63--1.12.} On the fixed $U=1.2$B ladder, the two-epoch point lies on the
fresh curve, and loss keeps falling through 16 epochs while separating
increasingly from it.

\paragraph{\arxnew{Fixed-budget sweeps.}}
At a fixed training-token budget, every \revise{model size} has higher loss with
$R$ epochs over $U=D/R$ than with one epoch over $D$ fresh tokens
(Figure~\ref{fig:study_design}d). At two epochs the excess is at the resolution
threshold (two independent one-epoch 59M runs differ by $0.0011\text{ bpb}$), so its order
across sizes is not interpretable. It then grows with each doubling of $R$ and, from $R=4$, with
model size. The excess stays below 2\% of the fresh loss through $R=4$ and reaches
10--50\% at $R=16$. At $R=32$, loss is 1.4 (31M) to 3.8 (472M) times its fresh
value; training loss continues to fall at 252M and 472M, indicating overfitting rather
than divergence. \arxlatest{Fitting Eq.~\eqref{eq:tpp_grid} with the size factor $N_{\rm total}^{-g}$ to this
sweep gives $p=0.88$ and $g=0.46$: the variable carries over, while its exponent and size dependence
vary with the setup.} Partial repetition does not escape this pattern. Across the 59M
five-source sweep ($D=2$B) and \arxlatest{CommonCrawl panels from 252M to 2B} ($D=8$B), no partial
allocation beats fresh data by more than $0.0023\text{ bpb}$, barely above the
resolution threshold. Drawing the whole budget from 32 epochs instead raises loss to
1.6--3.7 times its fresh value (Figure~\ref{fig:frcontour}).

\section{\texorpdfstring{\arxnew{Partial Repetition in the Subset's Repetition Variable}}{Partial Repetition in the Subset's Repetition Variable}}
\label{app:partial}
The 45M source study has nine allocations $(f,R_{\rm sub})$ on three sources at
$N=45{,}084{,}900$ ($N_{\rm total}=77{,}249{,}380$) and $D=805{,}306{,}368$. For each allocation we
define the repeated subset's variable
\[
z_{\rm sub}=(R_{\rm sub}-1)\,\frac{N_{\rm total}}{fD/R_{\rm sub}},
\]
which reduces to $z$ of Eq.~\eqref{eq:tpp_grid} at $f=1$, and fit the mean excess loss of the 26
source--allocation pairs above 0.002~bpb in $\log\Delta L$. Scaling the grid law by the replaced
fraction, $\Delta L=f\,A\,z_{\rm sub}^{p}$ with the grid's $A$ and $p$, \arxnew{overstates these pairs
by a factor of 3.9 on average (geometric mean), and with $A$ refitted it still leaves $R^2<0$}. A free fit
$\Delta L=c\,f^{a}z_{\rm sub}^{b}$ gives $a=1.63$ and $b=0.53$ ($R^2=0.89$; 0.84--0.97 when fitted per
source). At the same $z_{\rm sub}$, a larger repeated fraction therefore costs more than in proportion,
and \arxlatest{$z_{\rm sub}$ alone does not describe partial repetition; the repeated fraction enters separately}.

\begin{figure}[!htbp]\centering
\includegraphics[width=\linewidth]{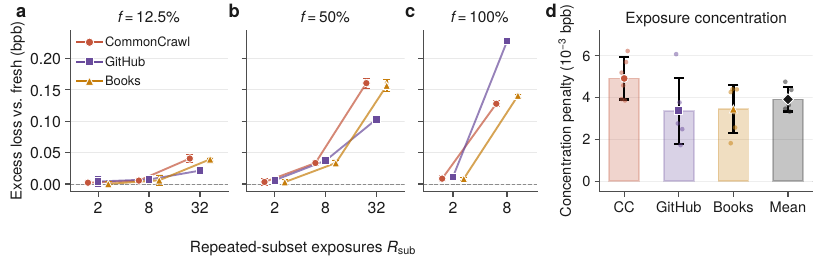}
\caption{\textbf{\arxlatest{Partial repetition raises mean loss (a--c)}, and matched
aggregate counts do not give matched loss (d).} Paired
study; 45M non-embedding parameters, \revise{$D\approx805$M}. (a--c)~Excess loss
relative to the same-source \revise{all-fresh run} per allocation $(f,R_{\rm sub})$, where $f$ denotes the fraction of training tokens drawn from the repeated subset:
lines join source means with pointwise 95\% $t$ intervals ($n=3$); dots
are paired replicates. (d)~\revise{Concentration penalty: concentrated \revnew{$(f=33.3\%, R_{\rm sub}=4)$} minus spread \revnew{$(f=50\%, R_{\rm sub}=2)$}. Both allocations have equal training-token, distinct-target and total-extra-exposure counts; the concentrated one repeats a subset one third as large and is not shown in (a--c).} Bars show per-source and \revise{equal-source} mean penalties with 95\% $t$ intervals over six paired replicates (points)\arxlatest{. All 18 source--replicate differences are positive, and a sign-flip test on the six replicate
means gives $p=0.031$, the smallest attainable.}}
\label{fig:replacement}\label{fig:matched_content}
\end{figure}

\FloatBarrier
\revnew{\section{Scale Studies at Fixed Budgets per Parameter}
\label{app:scale_new}\label{app:paired_scale}
Both studies use the paired-study architecture family and recipe (Appendix~\ref{app:protocols}):
width/depth ratio 64, head dimension 64, grouped-query attention with half as many KV heads,
ReLU$^2$ MLPs and tied embeddings, at seven exact non-embedding sizes from 23,087,168 (width 512,
depth 8) to 1,216,772,100 (width 1920, depth 30), trained and evaluated separately on CommonCrawl
and GitHub with the same development and test sets. Global batches grow from 131,072 to 2,097,152
targets with size. Momentum warms up over the first 10\% of updates and the learning rate decays
linearly over the final 20\%.

\paragraph{\arxlatest{Partial repetition at $D=20N$.}}
Five sizes from 45M to 1.2B train on $D=20N$ targets with fresh data, \revnew{$(f,R_{\rm sub})=(50\%,2)$} or
\revnew{$(50\%,8)$}, \arxlatest{with three paired replicates (Appendix~\ref{app:paired})}; the excess loss
is the difference from the same-replicate fresh run. The matrix learning rate is chosen per size
from fresh-data calibration runs on development loss (0.01 at 45M and 185M, 0.005 from 361M).

\paragraph{Full corpus at $U=2.5N$.}
Seven sizes from 23M to 1.2B train for $R\in\{1,2,4,6,8,12\}$ epochs over $U=2.5N$ unique
targets, so $D=2.5RN$ and $R=8$ matches the \arxlatest{partial-repetition} budget $D=20N$. \arxnew{As on the grid, each
epoch is an independent shuffle of the same $U$ targets.} Each run is separately
annealed\arxlatest{, with one replicate}. The learning rate is set per size and
source by a power law in compute fitted to the calibration optima, evaluated at $D=20N$ and shared
by all $R$ of a size, so no learning rate is tuned per $R$. $R^*$ in Figure~\ref{fig:scale}a is the
vertex of a least-squares quadratic in $\log R$ through all points of a size; the curves in
Figure~\ref{fig:scale}b,c are a separate six-parameter surface per source,
$\log L=\sum_{i\le1,\,j\le2}c_{ij}(\log N-\overline{\log N})^i(\log R)^j$, whose minima (diamonds)
differ from the per-size vertices by at most 0.4 epochs.

\input{tables/scale-studies-arxiv.tex}

\paragraph{\arxlatest{A two-size study.}}
\arxlatest{A paired study compares 45M and 185M models at matched $D/N$ ($D\approx4.47N$) and matched
$U/N$ ($U\approx1.12N$), with three replicates per source (Figure~\ref{fig:scale_paired}; decision rules
in Appendix~\ref{app:paired}). Its unique budgets, 5--20 times below the fixed-$U$ sweeps, also serve as
the out-of-range check of Appendix~\ref{app:posthoc}.}

\begin{figure}[!htbp]\centering
\includegraphics[width=\linewidth]{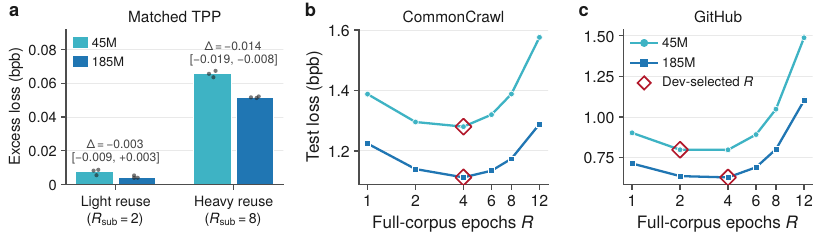}
\caption{\textbf{\arxlatest{At matched $D/N$, the 185M model pays less for heavy partial repetition; at matched
$U/N$, both sizes are best near four epochs.}}
Two sources, three paired replicates per source; exact non-embedding parameters. (a)~Matched $D/N$: \arxlatest{excess loss under partial repetition} for 45M vs. 185M models under light (\revnew{$f=50\%, R_{\rm sub}=2$}) and heavy (\revnew{$f=50\%, R_{\rm sub}=8$}) reuse. For heavy reuse, the larger-minus-smaller contrast is $-0.0138\text{ bpb}$, 21\% of the smaller configuration's penalty. (b,c)~Matched $U/N$: test loss of separately annealed full-corpus runs; lines join replicate
means. Diamonds mark the repeat count selected on development data.}
\label{fig:scale_paired}
\end{figure}

At matched $D/N$, the 185M penalty for heavy partial repetition (\revnew{$f=50\%, R_{\rm sub}=8$}) is
$0.0138\text{ bpb}$ smaller than at 45M (21\%; 95\% CI $[-0.0192,-0.0083]$), but does not vanish:
$0.069/0.062\text{ bpb}$ (CommonCrawl/GitHub) at 45M against $0.054/0.049\text{ bpb}$ at 185M, about
twice the penalties at $D=20N$ (Table~\ref{tab:scale_studies}). The contrast is already negative on
development loss after the first $20\%$ of training (Figure~\ref{fig:dynamics}c), so the final
learning-rate decay does not produce it alone. Light partial repetition \revnew{($f=50\%, R_{\rm sub}=2$)}
gives $-0.0030\text{ bpb}$ (95\% CI $[-0.0093,0.0032]$).

At matched $U/N$, mean test loss is lowest at four epochs in all four source--size
combinations (45M/GitHub: two epochs within $0.0006\text{ bpb}$), close to the minima at $U=2.5N$
(Figure~\ref{fig:scale}), \arxlatest{which extends this arm to seven sizes}.}

\FloatBarrier
\section{\texorpdfstring{\arxlatest{The Fixed-Unique-Data Sweeps}}{The Fixed-Unique-Data Sweeps}}
\label{app:model_checks}\label{app:datarecords}\label{sec:extrap}
\arxlatest{The surface fits below use 73 endpoints. The 511M/$R=4$ endpoint enters the loss minima,
Figure~\ref{fig:rpeak} and Tables~\ref{tab:fixed_u} and~\ref{tab:second_epoch}, but not these fits
(Appendix~\ref{app:repro}).}

The fixed-$U$ ladders use separately annealed runs at each horizon, not
intermediate checkpoints of one schedule. The 14M ladder rises by only 0.0010
bpb from 16 to 32 epochs, and the 31M and 59M ladders improve through 32\revise{; a
follow-up at 64 epochs finds no upturn for any of the three
(App.~\ref{app:r64})}. \arxnew{Each minimum is the vertex of a parabola in $\log R$ through the
lowest measured point and its two neighbors.} The 511M parabola uses its $R=4$ endpoint (Appendix~\ref{app:repro}). \arxlatest{Treating the ladders
without an upturn as lower bounds gives an exponent of 0.708.}

\paragraph{Same $U$ and $R$ across model sizes.}
Table~\ref{tab:fixed_u} gives each fixed-$U$ ladder relative to the same
model's one-epoch run at the same $U$, so every column compares sizes at an
identical $(U,R)$. At both unique budgets the second-epoch gain grows with size:
from 0.031 to 0.063--0.067 bpb between 14M and 511M--2B at $U=1$B, and from
0.022 to 0.036 bpb between 59M and 1B at $U=4$B. At $U=1$B only the 1B and 2B
models end above their one-epoch loss, from $R=16$ onward. At $U=4$B, where
$U/N\ge3.97$ for every size, all four ladders still improve at $R=8$, consistent
with the later minima that $R^\ast\propto(U/N)^{0.51}$ implies. \arxlatest{The reference here is the
data-matched baseline; Figure~\ref{fig:ds}d gives the compute-matched comparison at fixed $D$. The 511M
one-epoch endpoint lies above the 252M endpoint, which enlarges its relative gains.}
\begin{table}[!htbp]\centering\small
\caption{\textbf{Fixed-$U$ ladders relative to each model's own one-epoch run}
(CommonCrawl sweeps, CommonCrawl held-out bpb, $12Ld^2$ proxy sizes).
$L(1)$ is the one-epoch loss; other columns give $L(R)-L(1)$ in $10^{-3}$ bpb,
negative when the extra epochs help. The 511M/$R=4$ entry is described in Appendix~\ref{app:repro};
dashes mark horizons not trained.}
\label{tab:fixed_u}
\begin{tabular}{@{}lrrrrrr@{}}\toprule
$N$ & $L(1)$ & $R=2$ & $R=4$ & $R=8$ & $R=16$ & $R=32$\\\midrule
\multicolumn{7}{@{}l}{$U=1$B}\\
14M & 1.2029 & $-31.4$ & $-51.1$ & $-61.9$ & $-71.6$ & $-70.6$\\
31M & 1.1642 & $-36.6$ & $-61.2$ & $-76.1$ & $-85.8$ & $-90.9$\\
59M & 1.1343 & $-41.0$ & $-67.5$ & $-83.0$ & $-90.0$ & $-94.2$\\
127M & 1.1060 & $-45.9$ & $-74.0$ & $-88.2$ & $-91.1$ & $-87.3$\\
252M & 1.0837 & $-48.4$ & $-76.5$ & $-84.9$ & $-76.1$ & $-55.5$\\
511M & 1.0951 & $-67.4$ & $-101.1$ & $-102.3$ & $-70.1$ & $-8.6$\\
1B & 1.0696 & $-63.3$ & $-90.5$ & $-68.2$ & $+18.9$ & $+190.7$\\
2B & 1.0538 & $-65.3$ & $-90.0$ & $-44.7$ & $+120.2$ & $+496.0$\\\midrule
\multicolumn{7}{@{}l}{$U=4$B}\\
59M & 1.0605 & $-22.4$ & $-37.9$ & $-48.0$ & -- & --\\
252M & 0.9955 & $-28.3$ & $-46.7$ & $-59.2$ & -- & --\\
511M & 0.9788 & $-34.4$ & $-57.2$ & $-68.5$ & -- & --\\
1B & 0.9553 & $-35.8$ & $-57.0$ & $-64.8$ & -- & --\\\bottomrule
\end{tabular}
\end{table}

\revise{\paragraph{Second-epoch value across sizes and sources.}
Table~\ref{tab:fixed_u} measures gains relative to the data-matched baseline. To express them in fresh
tokens, we use the one-epoch CommonCrawl endpoints at $U=1$B and $U=4$B, which exist
for the 59M, 252M, 511M and 1B proxy sizes. For each size, we pass
$L=E+A\,U^{-a}$ through these two endpoints and invert it at the repeated endpoints
$L(1\text{B},R)$, which lie between them. The exponent $a=0.436$ is fitted to four
one-epoch 59M endpoints ($U=1,2,4,8$B; maximum residual $0.0005$ bpb) and shared
across sizes; bracketed ranges use $a/2$ and $2a$. Table~\ref{tab:second_epoch}
reports $\eta$ (Eq.~\ref{eq:ceg}) at $R=2$ and $R=4$, plotted in Figure~\ref{fig:rpeak}c. Second-epoch $\eta$ stays at
0.93--1.01 from 59M to 1B, whereas $\eta$ at four epochs falls from 0.81 to 0.59.
\arxlatest{This two-epoch level is close to the grid's (0.87--0.94 at $D/N\le50$).}
At 59M, measured one-epoch runs at $2U$ allow a direct check without interpolation.
At $U=1$, 2 and 4B, the second epoch recovers $98\%$, $95\%$ and $100\%$ of the loss
reduction from doubling fresh data. Across five sources at $U=1$B (CommonCrawl,
arXiv, Books, GitHub and Wikipedia), it recovers $98$--$101\%$. On the mixed corpus,
the fixed-budget sweep of Figure~\ref{fig:study_design}d is consistent: from 31M to
472M, two epochs over $D/2$ come within $0.0011$--$0.0029$ bpb of one epoch over $D$.
The $85\%$ estimate of \arxnew{Appendix~\ref{app:fixed_budget}}, at a far larger unique budget
per parameter ($U=25$B at 59M), is lower. \arxlatest{Run-to-run differences of 0.002--0.004~bpb correspond
to about 0.07 in second-epoch $\eta$.}}
\begin{table}[!htbp]\centering\small
\caption{\revise{\textbf{Value of a repeated token across proxy sizes}
(CommonCrawl, $U=1$B, $12Ld^2$ proxy sizes). $\eta$ inverts a
power law through each size's one-epoch endpoints $L_1$ at $U=1$B and $U=4$B;
brackets give the range for exponents $a/2$ and $2a$. The 511M/$R=4$ entry is described in
Appendix~\ref{app:repro}.}}
\label{tab:second_epoch}
\begin{tabular}{@{}lrrll@{}}\toprule
$N$ & $L_1(1\text{B})$ & $L_1(4\text{B})$ & $\eta$, $R=2$ & $\eta$, $R=4$\\\midrule
59M & 1.1343 & 1.0605 & 0.95 [0.76, 1.05] & 0.81 [0.75, 0.83]\\
252M & 1.0837 & 0.9955 & 0.93 [0.75, 1.03] & 0.72 [0.64, 0.75]\\
511M & 1.0951 & 0.9788 & 1.01 [0.82, 1.12] & 0.72 [0.65, 0.75]\\
1B & 1.0696 & 0.9553 & 0.94 [0.76, 1.05] & 0.59 [0.51, 0.63]\\\bottomrule
\end{tabular}
\end{table}

\begin{figure}[!htbp]\centering
\includegraphics[width=\linewidth]{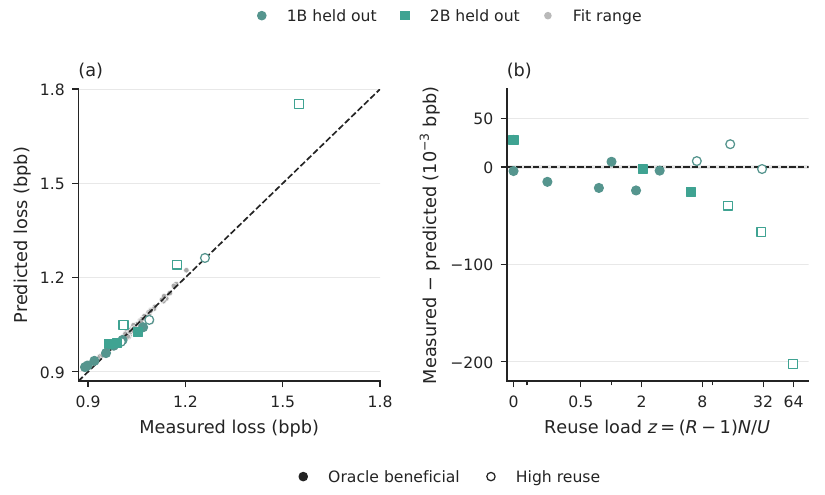}
\caption{\textbf{\arxlatest{Held-out-size check of the surface.}}
Eq.~\eqref{eq:mem} is fit to the 57 configurations with $N\le511$M and predicts
all 16 held-out 1B/2B endpoints. (a)~Measured versus predicted loss.
(b)~Measured-minus-predicted residuals versus $z$ on a symmetric-log axis that
includes the three $R=1$, $z=0$ points. Filled symbols mark the ten
\arxlatest{low-reuse points, including the three at $R=1$,} and hollow symbols the six high-reuse points.}
\label{fig:extrap}
\end{figure}

\begin{figure}[!htbp]\centering
\includegraphics[width=\linewidth]{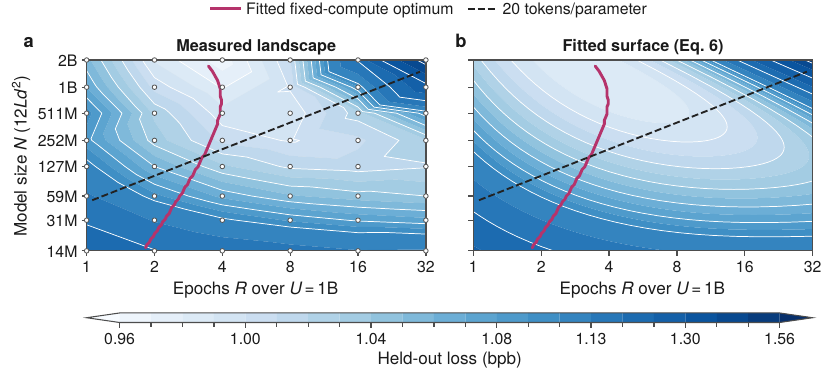}
\caption{\textbf{The count-based surface captures the shape of the measured
landscape at $U=1$B, but not its detail.}
(a)~Piecewise-linear interpolation of the measurements (circles) in log
coordinates. (b)~The 73-point surface of Eq.~\eqref{eq:mem}, evaluated at its
reported parameters on the same color scale and ending at the measured $R=32$
boundary. Solid curves minimize fitted loss along fixed compute $6NUR$ where the minimum is interior; the dashed line is a 20-tokens-per-parameter fresh reference.}
\label{fig:isoloss}
\end{figure}

\paragraph{\arxlatest{A count-based loss surface.}}\label{app:surface}
\arxlatest{We also fit the fixed-$U$ sweep endpoints with a surface that combines the saturating benefit of
\citet{muennighoff2023scaling} with the repetition term of \citet{lovelace2026prescriptive},}
\begin{equation}
L=E+\frac{A}{N^\alpha}+\frac{B}{D_{\rm eff}^{\beta}}+\lambda z^p,\quad
D_{\rm eff}=U\Bigl\{1+R_D\bigl[1-e^{-(R-1)/R_D}\bigr]\Bigr\},\quad
z=(R-1)\frac{N}{U},\label{eq:mem}
\end{equation}
\arxlatest{with $N$ the proxy count; saturation alone cannot produce a loss minimum at finite $R$
(Table~\ref{tab:forms}). Fitted up to 511M ($R^2=0.98$), the surface predicts the 1B and 2B endpoints
to 0.016~bpb \arxlatest{at low reuse and 0.057~bpb at high reuse} (Figures~\ref{fig:extrap}
and~\ref{fig:isoloss}). At unique budgets 5--20 times below its fitted range, however, it predicts that
12 epochs beat one, whereas all four measured source--size combinations get worse
(Appendix~\ref{app:posthoc}). Such surfaces are reliable only within the range they are fitted on, which
is why the main text reads the value of a repeat off measured curves.}

The fitted parameters are
\[
(E,A,\alpha,B,\beta,R_D,\lambda,p)=
(0.705,\,7.44,\,0.189,\,4.45\times10^{3},\,0.485,\,4.545,\,5.199\times10^{-3},\,1.160),
\]
with $N,U$ entered as raw proxy-parameter and token counts; other units change the coefficients.
\arxlatest{The surface's optimum depends on $U$ and $N/U$ separately. Its implied exponent of $R^\ast$ in $U/N$
is 0.410, against 0.509 from the five measured minima (Figure~\ref{fig:rstarcollapse}; leave-one-out
range 0.386--0.613).}

\arxnew{Figure~\ref{fig:isoloss} compares the 73-point surface with the measured landscape at $U=1$B.}

\input{tables/model-comparison.tex}

\arxlatest{Most residuals (63 of 73) exceed the 0.002-bpb resolution, so the surface captures the shape of the
landscape rather than its detail.}

\paragraph{Seen/unseen gap.}
\label{sec:reuse_gap}
In the CommonCrawl $U=1$B sweeps at 14M, 31M and 59M, the unseen-minus-seen
bpb gap increases with $R$, faster for larger models at matched $R$, and its
change precedes the held-out upturn. Across the 15 $R>1$ endpoints the
log--log correlation between $z^p$ and the gap is $r=0.956$; the three $R=1$
points ($z=0$) are omitted from that logarithmic comparison. The gap was not
used in the fit. Because multiplying $\log z$ by any positive exponent leaves
the correlation unchanged, the association does not identify $p$.

\section{\texorpdfstring{\arxlatest{Partial Repetition Sweeps, Sources and Order}}{Partial Repetition Sweeps, Sources and Order}}
\label{app:allocation}\label{sec:tolerance}
\begin{figure}[tbp]\centering
\includegraphics[width=0.72\linewidth]{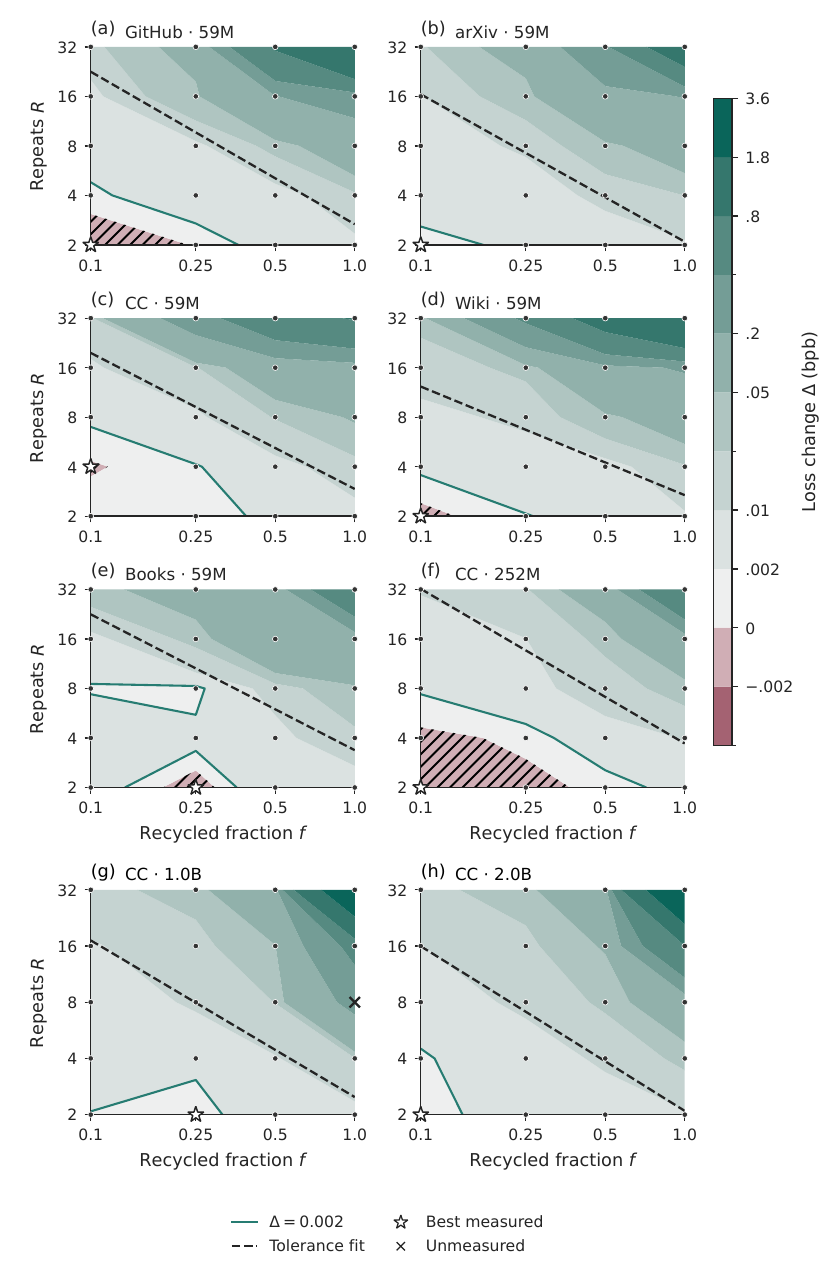}
\caption{\textbf{\arxlatest{Partial repetition} by domain and model size at fixed training-token budget
(sweeps).} Signed excess loss $L(f,R_{\rm sub})-L(0,1)$ relative to all-fresh training of the same
domain, model and budget; the axis $R$ denotes $R_{\rm sub}$. (a--e)~Five domains at 59M,
$D=2$B. (f--h)~CommonCrawl at 252M, 1B and 2B, $D=8$B; the 252M model uses four KV heads, unlike its
main-sweep counterpart. Dots are measured configurations; stars mark the lowest observed interior
loss; crosses mark unmeasured configurations, whose color is interpolated. The solid contour marks
0.002~bpb and dashed lines the per-panel 0.01-bpb tolerance fits\arxlatest{; contour intervals are nonlinear}.}
\label{fig:frcontour}
\end{figure}

\paragraph{\arxlatest{Partial-repetition sweeps.}}
\arxlatest{In the single-run sweeps of Figure~\ref{fig:frcontour}, loss rises with $R_{\rm sub}$ at almost every
repeated fraction: 21 of 24 columns are nondecreasing within the 0.002-bpb resolution, and the three
exceptions, all on Books, remain at least 0.115~bpb worse than fresh data.} Let $R_{\rm safe}(f)$ be the largest $R_{\rm sub}$ whose excess over
all-fresh training stays within a 0.01-bpb tolerance, log-interpolated along
each $f$ column; on CommonCrawl it follows $R_{\rm safe}\approx\kappa f^{-|b|}$
with $(\kappa,|b|)\approx(2.7,0.87)$. Changing the tolerance to 0.005 or 0.02 gives
$(\kappa,|b|)=(1.95,0.80)$ or $(3.93,1.09)$.
Across fractions and sizes, $fR_{\rm safe}$ ranges 1.65--4.07. \arxlatest{Over the 54 points with small
positive penalties, $\Delta\propto(fR)^{1.06}$ ($R^2_{\log}=0.68$, against 0.35 for $R$ alone). At fixed
$D=8$B (Figure~\ref{fig:frcontour}f--h), the largest-recycling penalty rises from 0.83 to 3.39~bpb between
252M and 2B, as fixed $D$ implies.}

\paragraph{\arxlatest{Source entropy.}}
\revise{The source-entropy sweep in Figure~\ref{fig:entropy} trains the 59M-proxy model (54.1M counted
non-embedding parameters) separately on seven SlimPajama sources with $U=0.2$B unique tokens each and
$R\in\{1,2,4,8,16\}$, for \arxnew{35} runs in total. The loss rise is
measured on each source's own held-out split, from its minimum over $R$ to $R=16$; the gap is
held-out-probe minus training-probe loss, as in Figure~\ref{fig:schedule}c. Both source statistics
use every eighth document: unigram entropy over its first 2,048 GPT-2 tokens, and the gzip ratio
(level 6) as compressed over raw bytes of its first 8~KiB of UTF-8 text, summed over documents. Exact
permutation $p$-values enumerate all $7!$ assignments of the statistic to sources.} \arxnew{For the
loss rise against unigram entropy ($r=-0.89$), $p=0.008$.}

\paragraph{\arxlatest{Order.}}
\arxlatest{The 15 order runs (Figure~\ref{fig:schedule}) share $(N,U,R)$ within each configuration and differ only
in the order of repeats. Consecutive replay shrinks the unseen-minus-seen gap from 0.10 to 0.005~bpb at
511M and $R=8$ while raising held-out loss.}

\begin{figure}[!htbp]\centering
\includegraphics[width=\linewidth]{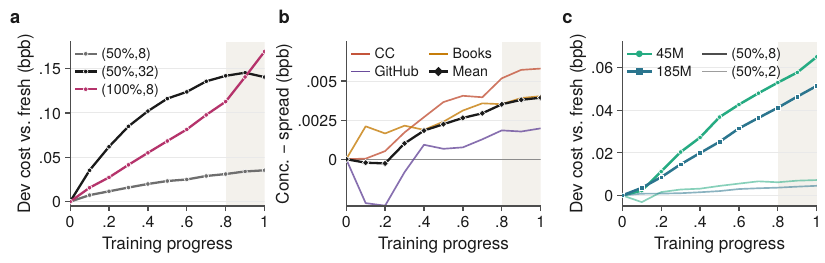}
\caption{\textbf{\arxlatest{The excess loss of partial repetition} builds up during training (development
loss).} Each point subtracts the same-source, same-replicate fresh run at the same
training progress; the learning rate decays linearly over the shaded final
20\%. (a)~\arxlatest{45M partial-repetition study}: mean over three sources and three replicates. (b)~\arxlatest{Concentration penalty} \revnew{$(33.3\%,4)$} minus \revnew{$(50\%,2)$} per source and
mean, three replicates. (c)~\arxlatest{Two-size study} at matched $D/N$ \revnew{($D\approx4.5N$)}, mean
over two sources and three replicates; lighter 45M, darker 185M.}
\label{fig:dynamics}
\end{figure}

\paragraph{\arxlatest{Early checkpoints can misrank allocations.}}
\arxlatest{The paired studies record development loss at every tenth of training, and
Figure~\ref{fig:dynamics} subtracts the same-replicate fresh run at each point. Intermediate points belong
to one run under its own schedule, not to separately annealed runs, and need not preserve the final
ranking. In the 45M study, \revnew{$(50\%,32)$} costs 0.116~bpb at the midpoint against 0.068 for
\revnew{$(100\%,8)$}; during the final learning-rate decay the full-corpus penalty grows from 0.113 to
0.169~bpb while the partial one stays near 0.14, so the two swap order only in the last tenth of training.
The concentration penalty is near zero for the first fifth of training and then rises steadily to 0.0039.
In the two-size study, the 185M-minus-45M \revnew{$(50\%,8)$} gap is negative from 20\% onward and ends at
$-0.0136$ (test $-0.0138$).}

\FloatBarrier
\section{\texorpdfstring{\arxlatest{Approximate IsoFLOP Profiles at Fixed Unique Data}}{Approximate IsoFLOP Profiles at Fixed Unique Data}}
\label{app:compute}
\begin{figure}[t]
\centering
\includegraphics[width=\linewidth]{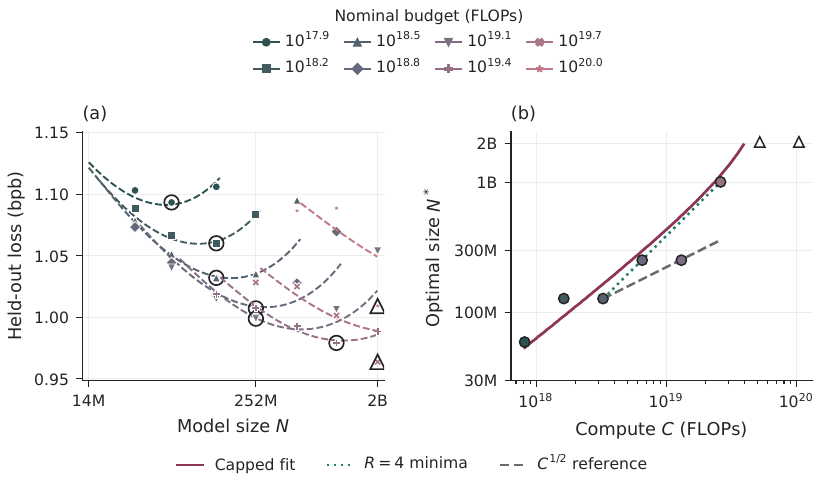}
\caption{\textbf{Approximate IsoFLOP profiles at $U=1$B.} Points are grouped by nearest half-octave in
$\log_2C$, with at least three per bin and up to $8.33\%$ actual-budget variation. (a) Retained
observations and fitted curves at bin-center compute. Outlined circles mark six interior grid minima and
triangles two boundary-limited minima. (b) Selected model size versus bin-center compute. The six-point
slope is $0.697$. The dotted fit uses the three interior minima at $R=4$ (slope
$0.995$); the dashed line is a slope-$1/2$ reference.}
\label{fig:isoflop}
\end{figure}

\arxlatest{The fixed-$U$ sweeps at $U=1$B are binned by compute into half-octaves, with up to 8.33\% budget
variation within a bin (Figure~\ref{fig:isoflop}). Six bins have an interior minimum and two are limited by
the largest model. Model size at the minimum grows as $C^{0.70}$ over the six interior minima and as
$C^{1.0}$ over the three at $R=4$; the count-based surface gives 0.87 and 0.93. These exponents describe
which model size wins at a given compute, a different question from the epoch count that minimizes loss
for a given model.}

\section{\texorpdfstring{\arxlatest{Checks of the Count-Based Surface}}{Checks of the Count-Based Surface}}
\label{app:posthoc}
\arxlatest{The two checks below evaluate the 73-point surface of Eq.~\eqref{eq:mem} with the parameters of
Appendix~\ref{app:model_checks} and the $12Ld^2$ size proxy, without refitting. The surface was fit to the
sweeps' byte-normalized metric, whereas the paired studies use ordinary-target bpb on their own holdouts,
so the comparisons indicate direction and scale rather than calibrated agreement.}

\paragraph{\arxlatest{Per-sample effective data and the concentration penalty.}}
Applying the saturation $s(r)=1+R_D[1-e^{-(r-1)/R_D}]$ to each 1,024-target
sample instead of at corpus level gives
$D_{\rm eff}=1024\,[196{,}608\,s(2)+393{,}216]=784.70$M targets for \revnew{$(50\%,2)$}
and $1024\,[65{,}536\,s(4)+524{,}288]=751.36$M for \revnew{$(33.3\%,4)$}. The benefit term
$B D_{\rm eff}^{-\beta}$ then predicts that \revnew{$(33.3\%,4)$} is worse by $+0.0046$
bpb, against the observed $+0.0039$\arxlatest{; the penalty term, defined by a full-corpus repeat
count, is omitted}. With the saturation
constant $R_D=15.4$ of \citet{muennighoff2023scaling} instead of the fitted
4.545, the prediction falls to $+0.0016$. \arxlatest{Sample-wise diminishing returns thus account for
the sign and order of magnitude of the concentration penalty, while its size depends on $R_D$.}

\paragraph{The surface outside its range.}
The \arxlatest{two-size study's} full-corpus design uses $U=50.3$M and 206.0M unique targets for
the 45M and 185M models (proxy sizes 49.2M and 201.3M), about 5--20 times
below the smallest unique budget in the fitted sweep (1B). Over continuous $R\in[1,32]$ the surface
places the optimum at $R^*\approx8.3$ and 6.3, with predicted improvements over one epoch of 0.38 and
0.17 bpb, and predicts that 12 epochs are still better than one by 0.37 and 0.14
bpb. The measured means are lowest at four epochs in all four strata, with
improvements of 0.086--0.112 bpb, and 12 epochs are worse than one epoch by 0.19
and 0.59 bpb at 45M (CommonCrawl/GitHub) and by 0.06 and 0.39 bpb at 185M.
\arxlatest{The surface correctly places the optimum earlier for the larger model, since its penalty is
identical at matched $N/U$ while the benefit term scales as $U^{-\beta}$, but it overstates the benefit of
continued reuse at small $U$ by a wide margin. This is why Appendix~\ref{app:surface} treats the surface
as a description within its fitted range.}

\FloatBarrier
\section{\texorpdfstring{\arxlatest{Additional Runs of the Fixed-Unique-Data Sweeps}}{Additional Runs of the Fixed-Unique-Data Sweeps}}
\label{app:repro}
\paragraph{\arxlatest{The 511M four-epoch endpoint.}}
\arxlatest{The 511M/$R=4$ endpoint of the $U=1$B CommonCrawl ladder was trained with the architecture,
schedule and seed of the other 511M runs; an earlier run at this point used a different architecture
(width 1536, depth 18, head dimension 96) and is not used.} Its CommonCrawl held-out loss is 0.9940 bpb, between the $R=2$ (1.0277)
and $R=8$ (0.9928) endpoints. \arxlatest{It enters Figures~\ref{fig:rpeak} and~\ref{fig:isoloss}a and
Tables~\ref{tab:fixed_u} and~\ref{tab:second_epoch}.} \arxlatest{The 511M parabola through $R=4,8,16$ places the minimum at 5.81 epochs, and the five
minima give $R^\ast=4.515(U/N)^{0.509}$ ($R^2_{\log}=0.938$). The surface of Eq.~\eqref{eq:mem} and the IsoFLOP profiles use the 73 endpoints
without it.}

\revise{\paragraph{\arxlatest{64-epoch extension.}}
\label{app:r64}
The 14M, 31M and 59M ladders of Figure~\ref{fig:rpeak} show no upturn by
$R=32$, so their minima lie outside the sweep. \arxlatest{We extended them to $R=64$, counting an upturn
only if $L(64)$ exceeds the ladder's best $R\le32$ loss by more than 0.002~bpb; the surface of
Eq.~\eqref{eq:mem} predicts rises of $+0.0025$, $+0.0064$ and $+0.0133$~bpb from $R=32$ to 64. Retraining
the 59M/$R=32$ endpoint reproduces its loss to 0.0003~bpb.}
Relative to the best $R\le32$ loss, $L(64)$ changes by $+0.0015$ (14M; best at
$R=16$), $-0.0032$ (31M) and $-0.0006$ bpb (59M), so none of the three turns
upward by 64 epochs and the 31M ladder still improves. The surface's $R=64$
predictions miss by $-0.008$, $+0.015$ and $+0.026$ bpb (prediction minus
measurement). These runs enter only Figure~\ref{fig:rpeak}; $R^\ast$, the power law and the surface are unchanged.}

\section{\texorpdfstring{\arxlatest{The Re-tokenization Study}}{The Re-tokenization Study}}
\label{app:bpe_methods}
\arxlatest{This study asks whether presenting repeated bytes as new token sequences changes the value of a
repeat (\S\ref{sec:bonus}). It trains 45M models on CommonCrawl for $R\in\{1,2,4,8,16\}$ epochs under four
tokenization schedules, with three paired replicates each, for 60 models in total.}

\paragraph{\arxlatest{Design.}}
\arxlatest{BPE-dropout \citep{provilkov2020bpe} skips each merge with probability $p=0.05$, so every occurrence
of a document is segmented afresh while its bytes are unchanged. The four schedules apply it never
(\emph{off}), throughout (\emph{on}), or for all but the final 20\% or 50\% of updates (\emph{20\%-off},
\emph{50\%-off}). All schedules share the same data order, batch of 131,072 targets, $K=768R$ updates and
learning-rate decay over the final 20\% of updates, so the 50\%-off switch does not coincide with a change
in learning rate. All four schedules share the same three replicates.}

\paragraph{\arxlatest{Tokenization and budgets.}}
\arxlatest{All schedules use the GPT-2 vocabulary and merge list (HF tokenizers 0.22.2, byte-level, no prefix
space or normalization), and evaluation always uses the deterministic segmentation of the original token
IDs. Re-segmented text is longer, so dropout raises the number of training targets per exposure by up to
5\% and the padded positions per update up to twofold (Table~\ref{tab:bpe_primary_budget}); each update
normalizes the summed loss by its number of valid targets.}

\input{tables/bpe-primary-budget-arxiv.tex}

\paragraph{\arxlatest{Results.}}
\arxlatest{The benefit of re-tokenization grows with reuse. At $R=16$, always-on lowers loss relative to
always-off by 0.154~bpb (95\% CI $[0.144,0.164]$). Averaged over horizons, switching dropout off for the
final 20\% of training improves on always-on by 0.0028~bpb (Table~\ref{tab:bpe_primary_budget}), but a
longer off period hurts at high reuse: 50\%-off is worse than 20\%-off by 0.013 and 0.055~bpb at $R=8$ and
16, and worse than always-on by 0.093~bpb at $R=16$. Its loss first falls after the switch, from 1.232 at
the midpoint to 1.225 at about 55\% of training, then rises to 1.364 at the end: the immediate gain from
deterministic segmentation does not last.}

\paragraph{\arxlatest{Statistics.}}
\arxlatest{The 60 models give 375 development and 60 test evaluations. Each primary contrast averages a
policy difference over the five horizons within a replicate (20\%-off minus on; 50\%-off minus 20\%-off)
and uses a paired $t$ interval with $t_{0.975,2}=4.303$. The per-horizon contrasts of
Figure~\ref{fig:bpe_policies} also carry simultaneous Bonferroni intervals, with critical values 16.688
over the 14 contrasts among off, on and 20\%-off, and 19.455 over the 19 that involve 50\%-off.}

%% file: tables/experiment-families-arxiv.tex
\begin{table}[!htbp]\centering\small
\caption{\textbf{\arxnew{The grid maps the landscape; the other designs test whether the map transfers.}}
\arxlatest{Sizes are $12Ld^2$ proxies for the fixed-budget, fixed-$U$, partial-repetition, order and source-entropy sweeps, and
exact non-embedding counts elsewhere. Paired studies share data order and initialization across the
allocations they compare.} CC: CommonCrawl. \revise{Mixed: all seven SlimPajama
subsets in their natural proportions. 5 sources: CC, GitHub, arXiv, Books and Wikipedia.}}
\label{tab:families}
\setlength{\tabcolsep}{4pt}
\begin{tabular}{@{}lllll@{}}
\toprule
Experiment & Models & Controlled & Varied & Sources\\
\midrule
\multicolumn{5}{@{}l}{\emph{\arxlatest{Single-run sweeps}}}\\
\arxnew{$D/N\times R$ grid} (\S\ref{sec:map}) & \arxnew{23M--361M} & \arxnew{$N$, $D$} & \arxnew{$D/N\le300$, $R\le16$ ($U=D/R$)} & \arxnew{CC}\\
Fixed $U$ (\S\ref{sec:fixed_u}) & 14M--2B & $U$ & $N,R\le32$; $R=64$ (14M--59M) & CC\\
\arxlatest{$U=2.5N$ (\S\ref{sec:fixed_pool})} & \revnew{23M--1.2B} & \revnew{$U/N=2.5$} & \revnew{$N$, $R\le12$} & \revnew{CC, GitHub}\\
\arxlatest{Order (\S\ref{sec:schedule})} & 59M--511M & $N$, $U$, $R$ & order & CC\\
\arxlatest{Source entropy (\S\ref{sec:source})} & \arxlatest{59M} & \arxlatest{$N$, $U$} & \arxlatest{source, $R\le16$} & \arxlatest{7 sources}\\
\arxlatest{Fixed budget (App.~\ref{app:fixed_budget})} & 31M--472M & $N$, $D$ & $R$ ($U=D/R$) & mixed\\
\arxlatest{Long run (App.~\ref{app:fixed_budget})} & 59M & $N$ & $U$; $R$ at 1.2B & mixed\\
\arxlatest{Partial repetition (App.~\ref{app:allocation})} & \arxlatest{59M--2B} & $N$, $D$ & $f$, $R_{\rm sub}$ & 5 sources; CC\\
\midrule
\multicolumn{5}{@{}l}{\emph{\arxlatest{Paired studies, three or more replicates}}}\\
\arxlatest{Partial repetition (\S\ref{sec:surface})} & 45M & $N$, $D$ & fresh + 9 allocations & CC, GitHub, Books\\
\arxlatest{$D=20N$ (\S\ref{sec:matched_tpp})} & \revnew{45M--1.2B} & \revnew{$D/N=20$} & \revnew{$N$; fresh, \revnew{$(50\%,2)$}, \revnew{$(50\%,8)$}} & \revnew{CC, GitHub}\\
\arxlatest{Two sizes (App.~\ref{app:paired_scale})} & 45/185M & \arxlatest{$D/N$ or $U/N$} & $N$; allocation or $R$ & CC, GitHub\\
\arxlatest{Re-tokenization (\S\ref{sec:bonus})} & 45M & $N$, bytes, order & \arxlatest{schedule}, $R\le16$ & CC\\
\bottomrule
\end{tabular}
\end{table}

%% file: tables/paired-configurations-arxiv.tex
\begin{table}[!htbp]\centering\small
\caption{Batch and update settings of the paired runs. $B$ is global targets/update; $G$ is
accumulation. Each GPU receives 16,384 targets per microbatch on eight GPUs.
LR order is matrix/scalar/tied embedding. Partial-repetition rows have fixed $D$;
full-corpus rows have fixed $U$ and $D=RU$, with $R\in\{1,2,4,6,8,12\}$.}
\label{tab:paired-config}
\begin{tabular}{@{}llrrr@{}}\toprule
Study/model & LR tuple & $B$ & $G$ & Fixed budget / updates\\\midrule
Source / 45M & 0.01/0.01/0.05 &131,072&1&$D=805{,}306{,}368$; $K=6{,}144$\\
$D/N$ / 45M & 0.02/0.02/0.05 &131,072&1&$D=201{,}326{,}592$; $K=1{,}536$\\
$D/N$ / 185M & 0.02/0.02/0.05 &524,288&4&$D=824{,}180{,}736$; $K=1{,}572$\\
$U/N$ / 45M & 0.02/0.02/0.05 &131,072&1&$U=50{,}331{,}648$; $K=384R$\\
$U/N$ / 185M & 0.02/0.02/0.05 &524,288&4&$U=206{,}045{,}184$; $K=393R$\\\bottomrule
\end{tabular}
\end{table}

%% file: tables/paired-summary-arxiv.tex
\begin{table}[!htbp]\centering\small
\caption{\textbf{\arxlatest{Summary of paired contrasts.}} \arxlatest{Concentration: $(33.3\%,4)$ minus $(50\%,2)$ at matched counts in the 45M study, six replicates ($df=5$). Two sizes: 185M minus 45M excess loss at matched $D/N$, three replicates ($df=2$).}}
\label{tab:paired-summary}
\begin{tabular}{@{}llcc@{}}\toprule
Contrast & Stratum & Mean (bpb) & 95\% CI\\\midrule
Concentration & Source mean & $+0.0039$ & $[+0.0033, +0.0045]$\\
& CommonCrawl & $+0.0049$ & $[+0.0039, +0.0059]$\\
& GitHub & $+0.0034$ & $[+0.0018, +0.0049]$\\
& Books & $+0.0034$ & $[+0.0023, +0.0046]$\\
Two sizes, $(50\%,8)$ & Source mean & $-0.0138$ & $[-0.0192, -0.0083]$\\
Two sizes, $(50\%,2)$ & Source mean & $-0.0030$ & $[-0.0093, +0.0032]$\\\bottomrule
\end{tabular}
\end{table}

%% file: tables/scale-studies-arxiv.tex
\begin{table}[!htbp]\centering\small
\caption{\textbf{Scale studies at fixed budgets per parameter.} Full corpus: epochs at the loss
minimum at $U=2.5N$ (least-squares quadratic in $\log R$ through all points of a size; one replicate).
Partial repetition: test excess loss over the same-replicate fresh run at $D=20N$, mean $\pm$ 95\% $t$
half-width over three paired replicates.
CC: CommonCrawl; GH: GitHub.}
\label{tab:scale_studies}
\setlength{\tabcolsep}{3.5pt}
\resizebox{\linewidth}{!}{%
\begin{tabular}{@{}lrrrcccc@{}}
\toprule
& & \multicolumn{2}{c}{Full corpus $R^*$} & \multicolumn{2}{c}{Partial $(50\%,8)$} & \multicolumn{2}{c}{Partial $(50\%,2)$}\\
\cmidrule(lr){3-4}\cmidrule(lr){5-6}\cmidrule(l){7-8}
Size & $N_{\rm nonemb}$ & CC & GH & CC & GH & CC & GH\\
\midrule
23M & 23,087,168 & \arxnew{3.72} & \arxnew{3.06} & -- & -- & -- & --\\
45M & 45,084,900 & \arxnew{3.79} & \arxnew{2.84} & 0.0322 $\pm$ 0.0021 & 0.0337 $\pm$ 0.0060 & 0.0038 $\pm$ 0.0033 & 0.0043 $\pm$ 0.0013\\
78M & 77,898,384 & \arxnew{4.03} & \arxnew{2.75} & -- & -- & -- & --\\
185M & 184,623,360 & \arxnew{4.22} & \arxnew{2.68} & 0.0249 $\pm$ 0.0012 & 0.0296 $\pm$ 0.0007 & 0.0024 $\pm$ 0.0023 & 0.0041 $\pm$ 0.0010\\
361M & 360,563,600 & \arxnew{4.30} & \arxnew{2.66} & 0.0245 $\pm$ 0.0012 & 0.0270 $\pm$ 0.0023 & 0.0027 $\pm$ 0.0014 & 0.0029 $\pm$ 0.0004\\
0.8B & 792,099,620 & \arxnew{4.42} & \arxnew{2.61} & 0.0221 $\pm$ 0.0008 & 0.0273 $\pm$ 0.0021 & 0.0022 $\pm$ 0.0010 & 0.0033 $\pm$ 0.0007\\
1.2B & 1,216,772,100 & \arxnew{4.22} & \arxnew{2.51} & 0.0216 $\pm$ 0.0008 & 0.0267 $\pm$ 0.0023 & 0.0024 $\pm$ 0.0005 & 0.0033 $\pm$ 0.0016\\
\bottomrule
\end{tabular}}
\end{table}

%% file: tables/model-comparison.tex
\begin{table}[!htbp]\centering\small
\caption{\textbf{Penalty forms on the 73 sweep endpoints.} $R^2$ is
in-sample; MAE is over the 16 held-out-size 1B/2B endpoints
(Figure~\ref{fig:extrap}). Parameter counts include the shared benefit terms;
diagnostics are in Appendix~\ref{app:model_checks}.}
\label{tab:forms}
\begin{tabular}{@{}llrrr@{}}\toprule
Form & Penalty & Parameters & $R^2$ & MAE (bpb)\\\midrule
Benefit-only & none & 6 & 0.360 & 0.0884\\
Separable & $\lambda(R-1)^p(N/U)^q$ & 9 & 0.985 & 0.1142\\
Load-augmented & $\lambda z^p$ & 8 & 0.982 & 0.0311\\\bottomrule
\end{tabular}
\end{table}

%% file: tables/bpe-primary-budget-arxiv.tex
\begin{table}[!htbp]\centering\small
\caption{\textbf{\arxlatest{Primary tokenization contrasts and actual budgets.}} \arxlatest{Contrasts average the paired policy difference over the five horizons within each replicate, with 95\% paired-$t$ CIs ($n=3$). Right: training targets $D$ and padded positions $P$ relative to the deterministic segmentation $E$, as ranges over replicates and horizons.}}
\label{tab:bpe_primary_budget}
\begin{tabular}{lrr}\toprule
Primary contrast & Mean bpb & 95\% CI \\
\midrule
20\%-off minus On & $-0.0028$ & $[-0.0042,-0.0014]$ \\
50\%-off minus 20\%-off & $+0.0122$ & $[+0.0114,+0.0129]$ \\
\bottomrule\end{tabular}
\quad
\begin{tabular}{lrr}\toprule
Policy & $\gamma=D/E$ & $\rho=P/E$ \\
\midrule
Off & 1.000 & 1.03--1.04 \\
On 0.05 & 1.049 & 2.00 \\
20\%-off & 1.039 & 1.80--1.81 \\
50\%-off & 1.024 & 1.51--1.52 \\
\bottomrule\end{tabular}
\end{table}